\documentclass[sigconf]{acmart}

\renewcommand\footnotetextcopyrightpermission[1]{} % removes footnote with conference information in first column
\usepackage{epstopdf} %converting to PDF
\usepackage{xcolor}
\usepackage{longtable}
\usepackage{multirow}
\usepackage{multicol}
\usepackage{booktabs}
\usepackage{colortbl}
\usepackage{hhline}
\usepackage{graphicx}
\usepackage{caption}

\newcommand{\chapternote}[1]{{%
  \let\thempfn\relax% Remove footnote number printing mechanism
  \footnotetext[0]{\emph{#1}}% Print footnote text
}}

\begin{document}

\title{What your Facebook Profile Picture Reveals about your Personality}
%\titlenote{Produces the permission block, and
%	copyright information}
%\subtitle{Extended Abstract}
%\subtitlenote{The full version of the author's guide is available as
%	\texttt{acmart.pdf} document}

\author{Cristina Segalin}
\affiliation{%
	\institution{California Institute of Technology}
%	\streetaddress{1200 E California Blvd}
%	\city{Pasadena}
%	\state{California, USA}
}
\email{segalinc@caltech.edu}

\author{Fabio Celli}
\affiliation{%
	\institution{University of Trento}
%	\streetaddress{Via Sommarive 5}
%	\city{Trento}
%	\state{Italy}
}
\email{fabio.celli@unitn.it}

\author{Luca Polonio}
\affiliation{%
	\institution{CIMEC - University of Trento}
%	\streetaddress{Corso Bettini 31}
%	\city{Rovereto}
%	\state{Italy}
}
\email{luca.polonio@unitn.it}

\author{Michal Kosinski}
\affiliation{%
	\institution{Stanford University}
%	\streetaddress{655 Knight Way}
%	\city{Stanford}
%	\state{California, USA}
}
\email{michalk@stanford.edu}

\author{David Stillwell}
\affiliation{%
	\institution{University of Cambridge}
%	\streetaddress{Trumpington Street}
%	\city{Cambridge}
%	\state{UK}
}
\email{cds617@cam.ac.uk}

\author{Nicu Sebe}
\affiliation{
	\institution{University of Trento}
%	\streetaddress{via Sommarive 9}
%	\city{Trento}
%	\state{Italy}
}
\email{sebe@disi.unitn.it}

\author{Marco Cristani}
\affiliation{
	\institution{University of Verona}
%	\streetaddress{Strada le Grazie 15}
%	\city{Verona}
%	\state{Italy}
}
\email{marco.cristani@univr.it}

\author{Bruno Lepri}
\affiliation{%
	\institution{FBK}
%	\streetaddress{Via Sommarive, 18}
%	\city{Trento}
%	\state{Italy}
}
\email{lepri@fbk.eu}

\renewcommand{\shortauthors}{C.Segalin et al.}

\begin{abstract}
%People spend a considerable amount of effort forming and managing impressions. Social network platforms such as Facebook provide new channels for this fundamental process. Psychologists have also shown the important role played by subjects' personality traits in the way they manage the images to convey in self-presentations. Hence, understanding subjects' behavior on social media could provide interesting insights on their personality. In the current paper, we investigate automatic personality recognition 
People spend considerable effort managing the impressions they give others. 
Social psychologists have shown that people manage these impressions differently depending upon their personality. Facebook and other social media provide a new forum for this fundamental process; hence, understanding people's behaviour on social media could provide interesting insights on their personality. In this paper we investigate automatic personality recognition from Facebook profile pictures. We analyze the effectiveness of four families of visual features and we discuss some human interpretable patterns that explain the personality traits of the individuals. For example, extroverts and agreeable individuals tend to have warm colored pictures and to exhibit many faces in their portraits, mirroring their inclination to socialize; while neurotic ones have a prevalence of pictures of indoor places. Then, we propose a classification approach to automatically recognize personality traits from these visual features. Finally, we compare the performance of our classification approach to the one obtained by human raters and we show that computer-based classifications are significantly more accurate than averaged human-based classifications for Extraversion and Neuroticism. 
\end{abstract}

% Please copy and paste the code instead of the example below. \begin{CCSXML}
%\begin{CCSXML}
%	<ccs2012>
%	<concept>
%	<concept_id>10002951.10003227</concept_id>
%	<concept_desc>Information systems~Information systems applications</concept_desc>
%	<concept_significance>300</concept_significance>
%	</concept>
%	<concept>
%	<concept_id>10010405.10010455.10010459</concept_id>
%	<concept_desc>Applied computing~Psychology</concept_desc>
%	<concept_significance>300</concept_significance>
%	</concept>
%	<concept>
%	<concept_id>10010405.10010455.10010461</concept_id>
%	<concept_desc>Applied computing~Sociology</concept_desc>
%	<concept_significance>300</concept_significance>
%	</concept>
%	<concept>
%	<concept_id>10010405.10010455</concept_id>
%	<concept_desc>Applied computing~Law, social and behavioral sciences</concept_desc>
%	<concept_significance>100</concept_significance>
%	</concept>
%	</ccs2012>
%\end{CCSXML}
%
%\ccsdesc[300]{Information systems~Information systems applications}
%\ccsdesc[300]{Applied computing~Psychology}
%\ccsdesc[300]{Applied computing~Sociology}
%\ccsdesc[100]{Applied computing~Law, social and behavioral sciences}
%% check if the css works for you wrt to the categories below here
%%CRISTINA: sapete i numeri delle categorie? nell sito della sottomisione non ci sono 
%%\category{H.4.m}{Information Systems Applications}{Miscellaneous}
%%\category{J.4}{Computer Applications}{Social and Behavioral Sciences, Sociology, Psychology}

%\terms{Algorithms, Experimentation, Theory}
\vspace{-.2cm}

%\medskip MM '17, October 23 - 27, 2017, Mountain View, California, USA

\maketitle
\section{Introduction}
\chapternote{MM'17 - October 23--27, 2017, Mountain View, CA, USA.}

People, being social animals, spend a considerable amount of effort forming and managing impressions, especially in the initial stage of social interactions~\cite{goffman1959}. Communication platforms such as Facebook, Twitter and Instagram, provide new channels for this fundamental process. For example, several studies reported that Facebook users engage in actively creating, maintaining and modifying an image of themselves by adjusting their profiles, including status updates and pictures, and also displaying their likes and dislikes~\cite{GonzalesHancock2008}. Hence, the Facebook profile page can be considered as a mediated representation of a Facebook user. Moreover, although users may be tempted to enhance their self-presentations, friends who are both offline and online keep Facebook users' self-presentations in check. Therefore, as shown by~\cite{vazire.gosling04} the online profile usually reflects the offline profile, although slightly enhanced. Social psychology has also shown the important role played by people's personality traits in the way they manage the impressions to convey in self-presentations~\cite{leary2011personality,rosenberg2011}. Thus, understanding people's behavior on social media could provide interesting insights on their personality.

Literature in personality psychology reported that users can make accurate personality impressions from the information displayed in social network user profiles~\cite{GoslingGaddisVazire2007}, and have investigated the specific features from user profiles and photos that are more useful to create personality impressions~\cite{EvansGoslingCarroll2008}. Specifically, estimations of Extraversion and Agreeableness were strongly related to profile picture friendliness while estimations of Neuroticism were related to profile picture unfriendliness. On the same line, Utz~\cite{Utz2010} associated user Extraversion with photo expressiveness. More recent works focused on the personality types associated to the production of selfies \cite{diefenbach2017selfie, sorokowski2015selfie, musil.al.selfie2017}, witch is often associated to narcissism, and to honesty in self-presentation \cite{hall2013self}. Several works have also used social media data for automatic personality recognition~\cite{golbeck2011predicting,kosinski.al13,kosinski2013private,quercia.al11a,youyou.al15}. Most of the personality computing literature analyzes Facebook profile characteristics (e.g. education, religion, marital status, number of political organizations the users belong to, etc.), textual content of statuses~\cite{schwartz2013personality}, ego-network structural characteristics~\cite{quercia2012facebook}, and preferences (e.g. Facebook Likes)~\cite{kosinski2013private}. However, as shown by~\cite{van2012effects} Facebook photos may have more impact on judgments of some traits (e.g. Extraversion) than textual self-disclosures.
In a precursory study, Celli \emph{et al.}~\cite{celli2014automatic} dealt with the task of personality recognition from Facebook profile pictures. In their work, they used a bag-of-visual-word representation of images in order to extract non interpretable visual features and they performed prediction of the personality without analyzing which of them are more influential for a particular trait. Another limitation of~\cite{celli2014automatic} was the size of the dataset of profile pictures, based only on 100 Facebook users.

Taking these previous findings as inspiration, we investigate the effectiveness of four different families of visual features: (i) Computational Aesthetics based features (CA), Pyramid Histogram Of visual Words based features (PHOW), Image Analysis TOol based features (IATO), and Convolutional Neural Networks based features (CNNs). The experiments are performed on the Facebook profile pictures collected in the \textit{myPersonality} corpus~\cite{kosinski2013private}, a significantly larger dataset compared with the one used by Celli \emph{et al.}~\cite{celli2014automatic}; specifically, we use 11,736 Facebook profile pictures, each one belonging to a different user. Finally, we focus on two personality traits, Neuroticism and Extraversion (which appear to be more relevant for revealing the personality of Facebook users), and we compare the accuracy of human and computer-based personality judgements.
% DAVID Why?If I'm a reviewer, I'll be wondering whether you tried agreeableness, openness, and conscientiousness... and they didn't work, so you didn't mention them. MARCO: solved
Interestingly, our results show that computer-based classifications are significantly more accurate than averaged human-based classifications for both traits. Our results provide some evidence in line with that recently obtained by Youyou \emph{et al.}~\cite{youyou.al15}, wherein computer-based predictions based on Facebook Likes are more accurate than those made by the participants' Facebook friends.

In summary, the main contributions of the paper are: 
(i) we exploit four families of visual features to represent the profile pictures in order to extract meaningful, relevant and interpretable visual patterns that correlate with the users' personality traits; (ii) we propose a classification approach to automatically recognize the personality from these visual features; and (iii) we compare the performance of our classification approach to the performance obtained by human raters on two specific personality traits, Neuroticism and Extraversion.\vspace{-.4cm}

\section{Related Work}\label{prev}
In the last years, the interest in automatic personality recognition has grown (see Vinciarelli and Mohammadi for a comprehensive survey~\cite{vinciarelli2014survey}) and several works have started exploiting the wealth of data made available by social media~\cite{quercia.al11a, golbeck.al11b, biel.gaticaperez13, kosinski.al13}, microphones and cameras~\cite{pianesi.al08, jayagopi2009modeling, mohammadi.vinciarelli12, 
lepri.al12}, and mobile phones~\cite{staiano.al12, chittaranjan.al13}.
Two works addressed the automatic recognition of personality traits from self-presentation videos~\cite{biel.gaticaperez13, batrinca2011please}. Biel \emph{et al.}~\cite{biel.gaticaperez13} used a dataset of 442 vlogs and asked external observers to rate vlogger's personality types; instead Batrinca \emph{et al.}~\cite{batrinca2011please} recorded video self-presentations of 89 subjects in a lab setting, asking them to complete a self-assessed personality test.

Personality traits of a user were found to have an influence on his/her choice of the profile picture in Facebook~\cite{wu2014facebook}, 
which suggests that profile pictures can be used to gauge users' personality types. Despite this, the recognition of personality from profile pictures is a relatively new and challenging task. Previous works predicted 
personality traits from preferred pictures in Flickr using Computational Aesthetics based features~\cite{segalin2016AffComp} and Convolutional Neural Networks based features~\cite{segalin2016CNN}. The experiments were performed using PsychoFlickr, a corpus of 60,000 pictures tagged as favorite by 300 ProFlickr users~\cite{Cristani:Flickr:ACMMM:2013}. For each user, the corpus includes 200 randomly selected favorite pictures and two personality assessments. The first assessment has been obtained by asking the users to self-assess their personality traits using a validated questionnaire (BFI ~\cite{rammstedt2007measuring}), while the second one has been obtained by asking 12 independent assessors to rate the traits of the user. The approach based on Convolutional Neural Networks outperformed the one using Computational Aesthetics based features, obtaining an average accuracy of 54\% for self-assessed personality traits and of 65\% for personality traits assessed by external observers.

Other previous works predicted personality traits from Instagram pictures~\cite{ferwerda2016} (with a Root Mean Square Error ranging from 0.66 for Conscientiousness to 0.95 for Neuroticism) and from profile pictures of 100 Facebook users, obtaining an average F1-measure of 67\%~\cite{celli2014automatic}. Both Ferwerda \emph{et al.} and Celli \emph{et al.} used a very limited number of users (113 and 100, respectively). Ferwerda \emph{et al.} used more images for each user (about 200) and a limited set of features, just based on color statistics and the presence of faces and people in images. Additionally, Ferwerda \emph{et al.} have different objectives from ours, being focused on investigating the relationship between the personality of Instagram users and the way they manipulate their pictures by using photo filters.

An important issue with personality prediction is the relationship between self-assessed and observer-perceived personality types. As already mentioned, Youyou \emph{et al.} provided evidence that personality predictions made by computers based on Facebook Likes are more accurate than those made by Facebook friends~\cite{youyou.al15}. Nevertheless the evaluation of perceived personality rating
%DAVID what do you mean by "the evaluation of perceived personality rating is evolving" ? MARCO: tried to solve it
by humans still remain an hot topic~\cite{joshi2014automatic}, with results obtained by Hall \emph{et al.}~\cite{hall.al13} indicating that observers could accurately estimate Extraversion, Agreeableness and Conscientiousness of unknown profile owners.
% We aim at investigating how well a machine can predict personality types from profile pictures and compare this to human judgements.
\vspace{-.2cm}

\section{Facebook Pictures and Personality Traits}\label{data}
We use a corpus of 11,736 Facebook profile pictures, each one belonging to a different user.
The dataset was retrieved from \textit{myPersonality} corpus~\cite{kosinski2013private}, a database collected from Facebook using an app~\cite{stillwell2012mypersonality} that measured several psychometric tests. In return for their scores, users could volunteer their Facebook profile data and Facebook activities (status messages, Likes, etc.), demographic and other information.
\begin{figure}[!ht]
    \centering
	\includegraphics[width=0.7\linewidth,height=0.35\textheight]{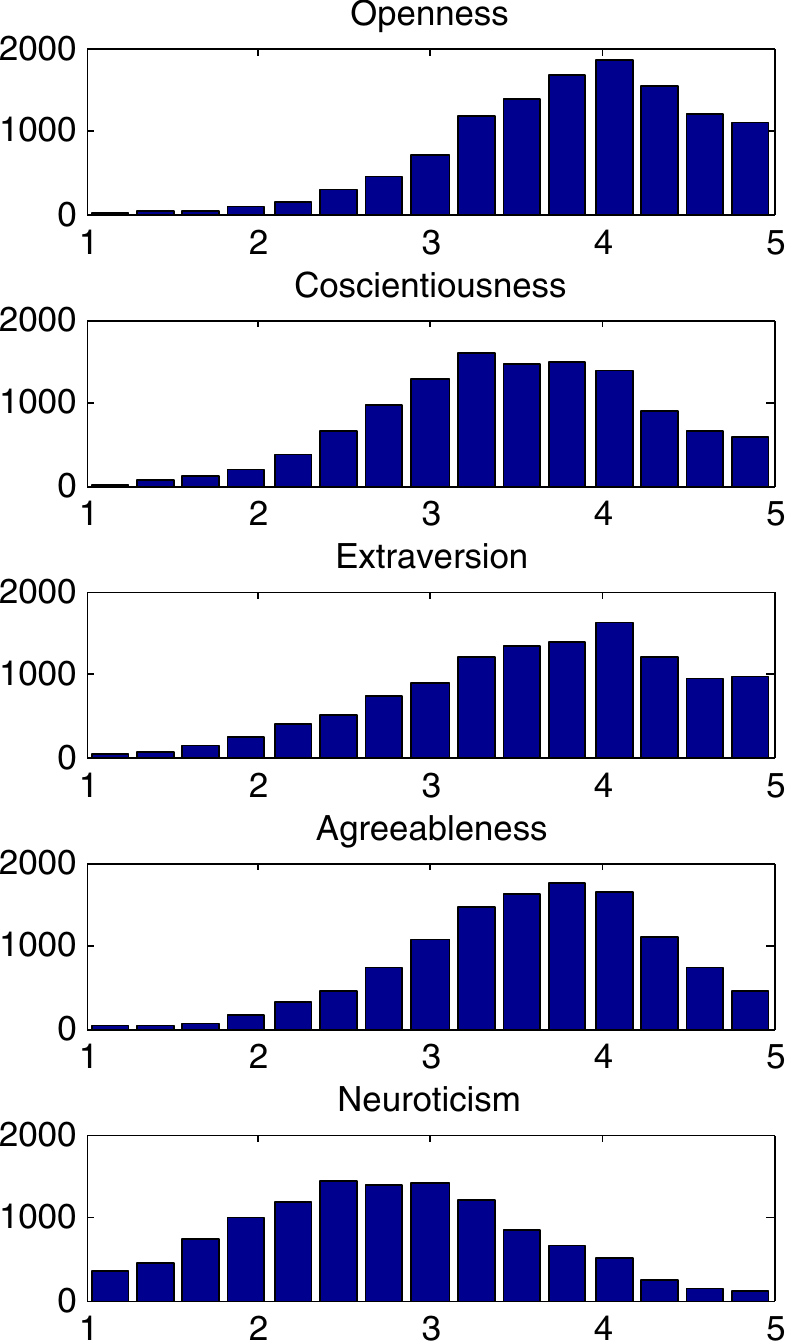}\vspace{-.2cm}
		\caption{Distribution of the collected self-assessments for each personality trait.}
		\label{fig:dist_trait}\vspace{-0.5cm}
\end{figure}
Self-assessment questionnaires have been used to measure and annotate the personality of each individual in terms of the \textit{Big Five} model~\cite{costa1985neo}. This model comprises five traits: (i) Extraversion (sociable, assertive vs. aloof, reserved), (ii) Agreeableness (friendly, cooperative vs. antagonistic, faultfinding), (iii) Conscientiousness (self-disciplined, organized vs. inefficient, careless), (iv) Neuroticism (insecure, anxious vs. calm, unemotional), and (v) Openness to Experience (intellectual, insightful vs. shallow, unimaginative). Personality was assessed using the International Personality Item Pool (IPIP) ~\cite{goldberg2006international}
questionnaire. The measurements consist of five scores, one per trait, that account for the position of an individual along each of the dimensions. Fig.~\ref{fig:dist_trait} reports the scores' distribution of each personality trait. We can notice that Openness to Experience, Conscientiousness, Extraversion, and Agreableness distributions are negatively skewed, while Neuroticism is positively skewed.%, an application for Facebook used for collecting psychological and socio-demographic dimensions of users with their consent.\\

\section{Feature Extraction}\label{feat}
\begin{table*}
	\centering
	\resizebox{1\textwidth}{0.22\textheight}{
		\renewcommand{\arraystretch}{1}
		{
			\begin{tabular}{|p{1cm}|p{2.7cm}|m{3.5cm}|c|m{12.5cm}|}
				%\begin{tabular}{|c|c|c|c|}
				\hline
				\textbf{Family} &\textbf{Category}  & \textbf{Name} & \textbf{d} & \textbf{Short Description} \\
				\hline
%--------------- AAVF
				% ------------------- color
				\multirow{16}{*}{CA}&\multirow{4}{*}{Color} & HSV statistics & 5 & Average of S channel and  standard deviation of S, V channels~\cite{Machajdik2012}; \textit{circular variance}  in HSV color space~\cite{mardia}; \textit{use of light} as the average pixel intensity of V channel~\cite{Datta06}  \\
				&& Emotion-based & 3 & Measurement of \textit{valence}, \textit{arousal}, \textit{dominance}~\cite{Machajdik2012}\\
				&& Color diversity & 1 &  Distance w.r.t a uniform color histogram, by Earth Mover's Distance (EMD)~\cite{Datta06, Machajdik2012} \\
				&& Color name & 11 &  Amount of \textit{black}, \textit{blue}, \textit{brown}, \textit{green}, \textit{gray}, \textit{orange}, \textit{pink}, \textit{purple}, \textit{red}, \textit{white}, \textit{yellow}~\cite{Machajdik2012} \\
				\cline{2-5}
				% -------------------- composition
				&\multirow{7}{*}{Composition} &  Edge pixels & 1 & Total number  of edge points, extracted with Canny detector~\cite{Lovato2012}\\
				&& Level of detail & 1 &  Number of regions (after mean shift segmentation)~\cite{Georgescu02}\\
				&& Average region size & 1 & Average \textit{size} of the regions (after mean shift segmentation)~\cite{Georgescu02} \\
				&& Low depth of field (DOF) & 3 & Amount of focus sharpness in the inner part of the image w.r.t. the overall focus~\cite{Datta06,Machajdik2012} \\
				&& Rule of thirds & 2 & Average of S,V channels over inner rectangle~\cite{Datta06,Machajdik2012}\\
				&& Image size & 1 & Size of the image~\cite{Datta06,Lovato2012}\\
				\cline{2-5}
				%----- texture--------------------------						
				&\multirow{4}{*}{Textural Properties} & Gray distribution entropy & 1 & Image entropy~\cite{Lovato2012} \\
				&& Wavelet based textures & 12 & Level of spatial graininess measured with a three-level (L1,L2,L3) Daubechies
				wavelet transform on  HSV channels~\cite{Datta06} \\
				&& Tamura & 3 & Amount of \textit{coarseness}, \textit{contrast}, \textit{directionality}~\cite{tamura} \\
				&& GLCM - features & 12 & Amount of \textit{contrast}, \textit{correlation}, \textit{energy}, \textit{homogeneousness} for each HSV channel~\cite{Machajdik2012}\\
				&& GIST descriptors& 24 &  Output of GIST filters for scene recognition~\cite{Oliva01}.\\
				\cline{2-5}
				% -------------------faces
				&\multirow{1}{*}{Content} & Faces & 1 &Number and size of faces after Viola-Jones face detection algorithm~\cite{Viola01}\\
				\hline
%------------------------- othe families
				PHOW & \multicolumn{2}{l|}{Pyramid Histogram Of visual Words} & 960 & A variant of the SIFT features extracted on a spatial pyramid~\cite{lowe2004distinctive}\\\hline
				CNN & \multicolumn{2}{l|}{Convolutional Neural Network} & 4096 & Features extracted at FC7 layer of the ImageNet network~\cite{krizhevsky2012imagenet}\\\hline
				\multirow{10}{*}{IATO} & \multirow{10}{*}{Image Analysis TOol} & Baseline scan info & 1 & Matching the byte sequence defining baseline scan \\
				& & Progressive scan info & 1 & Matching the byte sequence defining progressive scan\\
				& & Comments & 1 & Count of the byte sequence defining comments\\
				& & Huffman tables & 1 & Count of the byte sequence defining Huffman tables\\
				& & Quantization tables & 1 & Matching count of the byte sequence defining quantization tables  \\
				& & Start of image chars & 18 & 18 Features defining image characteristics: width, height, number of image components, and component subsampling\\
				& & Fillers & 1 & Count of x00 bytes\\
				& & Relative frequency of bytes & 255 & Relative frequency of each byte from x01 to xFF\\
				& & Total number of bytes & 1 & Size of the image in bytes\\ 
				\hline
				\end{tabular}}}
				\caption{Synopsis of the features. Every image is represented with 82 AAVF features split in four major categories (Color, Composition, Textural Properties, and Faces), 960 PHOW features, 4096 CNN features, 280 IATO features.}\label{table:feat}\vspace{-.4cm}
\end{table*}

We extract four families of features, ranging from easily interpretable ones, like Computational Aesthetics based features (CA), used in similar task as predicting personality from images liked by Flickr users~\cite{segalin2016AffComp}, to hardly interpretable ones, like Image Analysis TOol based features (IATO) and Pyramid Histogram Of visual Words based features (PHOW)~\cite{bosch2007image} that capture low level information and Convolutional Neural Network based features (CNN)~\cite{jia2014caffe} as they have been shown to be very performant on computer vision tasks. A description of each feature family follows:

%In the following, each family is briefly discussed, suggesting the reader to refer to the above papers for a more comprehensive description.
%\begin{description}
\textbf{CA - \textit{Computational Aesthetics based features}}: The cues of this family describe aesthetically, easily interpretable aspects of an image, like the use of the color, the localized presence of edges, the number of regions and their layout, etc. CA-based features have been recently used in~\cite{segalin2016AffComp} for linking the images liked by a person with his/her personality, and are organized into four main categories (color, composition, textual properties, and content), following the taxonomy proposed in~\cite{Machajdik2012}: a short description of their nature is reported in Table~\ref{table:feat}.

\textbf{PHOW - \textit{Pyramid Histogram Of visual Words based features}}: PHOW-based features~\cite{bosch2007image} are essentially (color) SIFT features~\cite{lowe2004distinctive} densely extracted at multiple scales in localized sectors of the images, and quantized into a bag of words representation~\cite{csurka2004visual}. These features have been considered for their massive usage in object recognition. In this paper, 3 scales are taken into account, with $4\times4$ sectors and a vocabulary of 20 words, resulting in a 960-dimensional feature vector. We tried different PHOW configurations before ending to this final one. In particular, we observed that if we increase the number of words some of them result redundant, while decreasing that number results in an underestimated and less representative number of words for our task. For the number of scales and sectors, even if we tried others configurations, we found that the standard values used by Vedaldi~\cite{Vedaldi07anopen} were adequate for our task.

\textbf{CNN - \textit{Convolutional Neural Network based features}}: Deep Learning tools such as Convolutional Neural Networks (CNNs) ~\cite{lecun1995convolutional} have recently become popular as methods for learning image representations~\cite{krizhevsky2012imagenet}. The usual pipeline for using CNNs is to exploit pre-learned network layers acting as filter banks on unseen data, or fine tuning the deepest layers on desired training data. In all the cases, the CNN training on images needs huge amount of samples (on the order of hundred of thousands in many cases), which is not our case. In this paper, we use the recently popular `ImageNet network'~\cite{krizhevsky2012imagenet} trained on 1.3 million images for the ImageNet~\cite{deng2009imagenet} challenge 2012. Specifically, we use Caffe~\cite{jia2014caffe} to extract features from the layer just before the final 1000 class classification layer, resulting in a feature vector of 4096 dimensions. These features provide a representation of each image at different levels of abstraction by applying different types of filters. 

\textbf{IATO - \textit{Image Analysis TOol based features}}: IATO is a new low-level feature extractor acting on the relative frequency of each byte of the picture, plus some specific format encoding characteristics of .jpeg, like Huffman tables and quantization tables. Table~\ref{table:feat} reports a description of the types and number of features that IATO can extract, resulting in a feature vector of 280 dimensions. Compared to the other feature sets, IATO is very fast and does not require high computational power. 

\section{Experiments}\label{exp}
The analyses carried out in the experiments are aimed to investigate two main research questions:
\begin{enumerate}
\item \textbf{Research question 1 (RQ1):} Do profile pictures contain information about their owners' personality?
\item \textbf{Research question 2 (RQ2):} Which, if any, of the profile picture features are indicative of their owners' personality traits?
\end{enumerate}
In the following, we will refer to RQ1 and RQ2 for addressing the specific research questions. This study is organized in two tasks: in the first one we perform a correlation analysis between image features and personality scores, in the second one we employ learning approaches to infer personality traits, given the features extracted from profile pictures. 

\subsection{Correlational Analysis}
Table~\ref{table:corr_en_num} presents the means of the absolute values of the statistically significant Spearman correlation coefficients between the features of a given family and the personality scores for each trait. The correlations' p-values are corrected using Bonferroni correction. The table reports the results obtained considering (i) all the samples below and above the mean $\mu$ of the personality scores' distributions and (ii) only the samples below the first quartile and above the third quartile $Q_{1,3}$, in order to account for strongly different profiles only. In addition, the table reports also the number of total features for each family (e.g. 82 for CA, 960 for PHOW, etc.), the number of features that correlate significantly with a trait (e.g. 61 using CA with Openness to Experience, 50 using CA with Conscientiousness, etc.) and their percentage with respect to the total number (e.g. 74\% using CA with Openness to Experience, 61\% using CA with Conscientiousness, etc.).
\begin{table*}[!ht]
	\centering
	\resizebox{\textwidth}{!}{%
		\renewcommand{\arraystretch}{1.1}
		{\normalsize
			\begin{tabular}{cc|c|c|c|c|c|c|c|c|c|c|}
				\cline{3-12}
				&  & \multicolumn{5}{c|}{Spearman $\rho$} & \multicolumn{5}{c|}{\# (\%) of correlations } \\ \cline{3-12}
				&  & O & C & E & A & N & O & C & E & A & N \\ \hline
				\multicolumn{1}{|c|}{\multirow{2}{*}{CA (82)}} & $\mu$ & 0.0168 & 0.0172 & 0.0174 & 0.0131 & 0.0163 & \underline{\textbf{61 (74\%)}} & 50 (61\%)& 59 (72\%)& \underline{\textbf{39 (47\%)}}& \underline{\textbf{69 (84\%)}} \\ \cline{2-12}
				\multicolumn{1}{|c|}{} & $Q_{1,3}$ & \underline{\textbf{0.0214}} & \underline{\textbf{0.0191}} & \underline{\textbf{0.0289}} & \underline{\textbf{0.0196}} & \underline{\textbf{0.0225}} & 57 (70\%) & \underline{\textbf{60 (73\%)}}& \underline{\textbf{62 (76\%)}}& 55 (67\%)& 65 (79\%)\\ \hline
				\multicolumn{1}{|c|}{\multirow{2}{*}{PHOW (960)}} & $\mu$ & 0.0150 & 0.0136 & 0.0155 & 0.0137 & 0.0140 & 641 (67\%)& 591 (62\%)& 631 (66\%)& 569 (59\%)& 586 (61\%)\\ \cline{2-12}
				\multicolumn{1}{|c|}{} & $Q_{1,3}$ & 0.0204 & 0.0179 & 0.0210 & 0.0187 & 0.0184 & 625 (65\%)& 589 (61\%)& 613 (64\%)& 605 (63\%)& 581 (61\%)\\ \hline
				\multicolumn{1}{|c|}{\multirow{2}{*}{CNN (4096)}} & $\mu$ & 0.0118 & 0.0119 & 0.0119 & 0.0116 & 0.0117 & 2040 (50\%)& 2100 (51\%)& 2031 (50\%)& 2011 (49\%)& 2032 (50\%) \\ \cline{2-12}
				\multicolumn{1}{|c|}{} & $Q_{1,3}$ & 0.0154 & 0.0155 & 0.0166 & 0.0152 & 0.0152  & 2052 (50\%)& 2064 (50\%)& 2057 (50\%)& 2032 (50\%)& 2006 (49\%)\\ \hline
				\multicolumn{1}{|c|}{\multirow{2}{*}{IATO (280)}} & $\mu$ & 0.0125 & 0.0120 & 0.0154 & 0.0121 & 0.0137 & 156 (57\%)& 155 (56\%)& 185 (66\%)& 163 (59\%)& 168 (61\%)\\ \cline{2-12}
				\multicolumn{1}{|c|}{} & $Q_{1,3}$ & 0.0163 & 0.0151 & 0.0227 & 0.0167 & 0.0190 & 156 (57\%)& 134 (49\%)& 177 (64\%)& 148 (54\%)& 193 (70\%)\\ \hline
			\end{tabular}
		}}
		\caption{Summary of the average of the absolute values of the statistically significant correlation scores of each feature family with all the traits (left) and number (and percentage) of the features which correlate significantly. \label{table:corr_en_num}}\vspace{-.6cm}
	\end{table*}

The results in the table trigger many interesting observations for RQ1: the amount of significant correlations found is considerable, which is encouraging and supports the previous findings that personality traits have an influence on the choice of profile pictures in Facebook~\cite{wu2014facebook}. Despite this, we observed that the correlation values are very low. 
%first of all, all the correlation values are very low, leaving open the doubts regarding the suitability of the features (RQ2) or the more crucial effective existence of a relation between self-portraits and personality (RQ1). Second, looking at the number of significant correlations found, one can observe that the number is indeed considerable,  for RQ1 %[BRUNO si puo' dire qualcosa sul numero, che tipo e' superiore al 5\% e quindi non c'e' roba a caso?].
However, the fact that the highest correlation values for all the traits and the highest percentage of correlated features is related to the CA-based features indicates that \textit{interpretability} is a key to select the appropriate cues to analyze the images (RQ2), and this motivates a further analysis that we will present in the following on the CA-based features. PHOW-based and IATO-based features are comparable in terms of correlation strength and percentage of significant correlations. CNN-based features are systematically a little lower as for the correlation strength and for the percentage, even if they represent as absolute number the highest amount of correlated cues. 
In Table~\ref{table:corr} we analyze in details the correlations between the CA-based features and the personality trait scores, considering only the samples above the first quartile and below the third quartile. For the sake of clarity and space, we report only those features that correlate significantly with at least two personality traits, reporting also those correlations that are significant at $p<0.01$. The first observations are that Extraversion and Neuroticism are the traits (in descending order) with more significant correlations. Considering the feature as independent variable, the number of faces in the image correlates with four traits (negatively with Openness to Experience, while positively with Conscientiousness, Extraversion, and Agreeableness); color features have three features that correlate each one with three traits (pink with Openness to Experience, Extraversion and Agreeableness, and red and yellow with Openness to Experience, Conscientiousness, and Extraversion).
\begin{table*}[!ht]
	\centering
	\resizebox{0.8\textwidth}{0.28\textheight}{
		\renewcommand{\arraystretch}{1.1}
		{\normalsize
			\begin{tabular}{|m{2.7cm}|p{5cm}|c|c|c|c|c|}
				\hline
				\textbf{Category}  & \textbf{Feature} & \textbf{Ope} & \textbf{Con} & \textbf{Ext} & \textbf{Agr} & \textbf{Neu} \\ \hline
				\multirow{11}{*}{Color}
			     & valence  & - & - & \cellcolor[rgb]{  0.80,  0.18,  0.18}\textcolor{white}{\textbf{  0.04}*}  & \cellcolor[rgb]{  0.80,  0.18,  0.18}\textcolor{white}{\textbf{  0.03}}  & - \\ \hhline{*{1}{|~}*{6}{|-}}
				 & colorfulness  & - & - & \cellcolor[rgb]{  0.16,  0.16,  0.80}\textcolor{white}{\textbf{ -0.04}*}  & - & \cellcolor[rgb]{  0.80,  0.18,  0.18}\textcolor{white}{\textbf{  0.02}}  \\ \hhline{*{1}{|~}*{6}{|-}}
				 & brown  & \cellcolor[rgb]{  0.16,  0.16,  0.80}\textcolor{white}{\textbf{ -0.03}*}  & \cellcolor[rgb]{  0.80,  0.18,  0.18}\textcolor{white}{\textbf{  0.03}}  & - & - & - \\ \hhline{*{1}{|~}*{6}{|-}}
				& pink  & \cellcolor[rgb]{  0.16,  0.16,  0.80}\textcolor{white}{\textbf{ -0.04}*}  & - & \cellcolor[rgb]{  0.80,  0.18,  0.18}\textcolor{white}{\textbf{  0.08}*}  & \cellcolor[rgb]{  0.80,  0.18,  0.18}\textcolor{white}{\textbf{  0.03}}  & - \\ \hhline{*{1}{|~}*{6}{|-}}
				& purple  & \cellcolor[rgb]{  0.16,  0.16,  0.80}\textcolor{white}{\textbf{ -0.03}*}  & - & \cellcolor[rgb]{  0.80,  0.18,  0.18}\textcolor{white}{\textbf{  0.07}*}  & - & - \\ \hhline{*{1}{|~}*{6}{|-}}
				 & red  & \cellcolor[rgb]{  0.16,  0.16,  0.80}\textcolor{white}{\textbf{ -0.03}*}  & \cellcolor[rgb]{  0.80,  0.18,  0.18}\textcolor{white}{\textbf{  0.02}}  & \cellcolor[rgb]{  0.80,  0.18,  0.18}\textcolor{white}{\textbf{  0.09}*}  & - & - \\ \hhline{*{1}{|~}*{6}{|-}}
				 & yellow  & \cellcolor[rgb]{  0.16,  0.16,  0.80}\textcolor{white}{\textbf{ -0.02}}  & \cellcolor[rgb]{  0.16,  0.16,  0.80}\textcolor{white}{\textbf{ -0.03}}  & \cellcolor[rgb]{  0.80,  0.18,  0.18}\textcolor{white}{\textbf{  0.04}*}  & - & - \\ \hline
				\multirow{4}{*}{Composition}
				 & pers. perc. of edges pixels  & - & - & \cellcolor[rgb]{  0.80,  0.18,  0.18}\textcolor{white}{\textbf{  0.03}*}  & \cellcolor[rgb]{  0.80,  0.18,  0.18}\textcolor{white}{\textbf{  0.03}}  & - \\ \hhline{*{1}{|~}*{6}{|-}}
				 & level of detail  & - & - & \cellcolor[rgb]{  0.80,  0.18,  0.18}\textcolor{white}{\textbf{  0.07}*}  & \cellcolor[rgb]{  0.80,  0.18,  0.18}\textcolor{white}{\textbf{  0.03}}  & \cellcolor[rgb]{  0.16,  0.16,  0.80}\textcolor{white}{\textbf{ -0.02}}  \\ \hhline{*{1}{|~}*{6}{|-}}
			     & avg region size  & - & - & \cellcolor[rgb]{  0.16,  0.16,  0.80}\textcolor{white}{\textbf{ -0.07}*}  & \cellcolor[rgb]{  0.16,  0.16,  0.80}\textcolor{white}{\textbf{ -0.03}}  & \cellcolor[rgb]{  0.80,  0.18,  0.18}\textcolor{white}{\textbf{  0.02}}  \\ \hhline{*{1}{|~}*{6}{|-}}
				& rule of thirds - saturation  & - & - & \cellcolor[rgb]{  0.80,  0.18,  0.18}\textcolor{white}{\textbf{  0.04}*}  & \cellcolor[rgb]{  0.80,  0.18,  0.18}\textcolor{white}{\textbf{  0.02}}  & - \\ \hline
				\multirow{23}{*}{Textural Properties}
				 & gray distribution entropy  & - & - & \cellcolor[rgb]{  0.80,  0.18,  0.18}\textcolor{white}{\textbf{  0.03}}  & \cellcolor[rgb]{  0.80,  0.18,  0.18}\textcolor{white}{\textbf{  0.03}}  & - \\ \hhline{*{1}{|~}*{6}{|-}}
				 & brightness wavelet - lev 2  & - & - & \cellcolor[rgb]{  0.80,  0.18,  0.18}\textcolor{white}{\textbf{  0.03}}  & - & \cellcolor[rgb]{  0.16,  0.16,  0.80}\textcolor{white}{\textbf{ -0.03}}  \\ \hhline{*{1}{|~}*{6}{|-}}
				 & brightness wavelet - lev 3  & - & - & \cellcolor[rgb]{  0.80,  0.18,  0.18}\textcolor{white}{\textbf{  0.04}*}  & - & \cellcolor[rgb]{  0.16,  0.16,  0.80}\textcolor{white}{\textbf{ -0.02}}  \\ \hhline{*{1}{|~}*{6}{|-}}
				 & brightness wavelet avg  & - & - & \cellcolor[rgb]{  0.80,  0.18,  0.18}\textcolor{white}{\textbf{  0.04}*}  & - & \cellcolor[rgb]{  0.16,  0.16,  0.80}\textcolor{white}{\textbf{ -0.03}}  \\ \hhline{*{1}{|~}*{6}{|-}}
				 & Tamura directionality  & - & - & \cellcolor[rgb]{  0.16,  0.16,  0.80}\textcolor{white}{\textbf{ -0.03}}  & \cellcolor[rgb]{  0.16,  0.16,  0.80}\textcolor{white}{\textbf{ -0.02}}  & - \\ \hhline{*{1}{|~}*{6}{|-}}
				& GLCM correlation - hue  & \cellcolor[rgb]{  0.16,  0.16,  0.80}\textcolor{white}{\textbf{ -0.03}*}  & - & \cellcolor[rgb]{  0.80,  0.18,  0.18}\textcolor{white}{\textbf{  0.03}}  & - & - \\ \hhline{*{1}{|~}*{6}{|-}}
				 & GLCM energy - saturation  & - & - & \cellcolor[rgb]{  0.16,  0.16,  0.80}\textcolor{white}{\textbf{ -0.06}*}  & - & \cellcolor[rgb]{  0.80,  0.18,  0.18}\textcolor{white}{\textbf{  0.03}*}  \\ \hhline{*{1}{|~}*{6}{|-}}
				 & GLCM homogeneity - saturation  & - & - & \cellcolor[rgb]{  0.16,  0.16,  0.80}\textcolor{white}{\textbf{ -0.04}*}  & - & \cellcolor[rgb]{  0.80,  0.18,  0.18}\textcolor{white}{\textbf{  0.02}}  \\ \hhline{*{1}{|~}*{6}{|-}}
				 & GLCM contrast - brightness  & - & - & \cellcolor[rgb]{  0.80,  0.18,  0.18}\textcolor{white}{\textbf{  0.04}*}  & - & \cellcolor[rgb]{  0.16,  0.16,  0.80}\textcolor{white}{\textbf{ -0.03}}  \\ \hhline{*{1}{|~}*{6}{|-}}
				 & GLCM energy - brightness  & - & \cellcolor[rgb]{  0.16,  0.16,  0.80}\textcolor{white}{\textbf{ -0.03}}  & \cellcolor[rgb]{  0.16,  0.16,  0.80}\textcolor{white}{\textbf{ -0.04}*}  & - & - \\ \hhline{*{1}{|~}*{6}{|-}}
				 & GLCM homogeneity - brightness  & - & - & \cellcolor[rgb]{  0.16,  0.16,  0.80}\textcolor{white}{\textbf{ -0.04}*}  & - & \cellcolor[rgb]{  0.80,  0.18,  0.18}\textcolor{white}{\textbf{  0.03}}  \\ \hhline{*{1}{|~}*{6}{|-}}
				 & GIST - channel 7  & - & - & \cellcolor[rgb]{  0.80,  0.18,  0.18}\textcolor{white}{\textbf{  0.03}}  & - & \cellcolor[rgb]{  0.16,  0.16,  0.80}\textcolor{white}{\textbf{ -0.03}*}  \\ \hhline{*{1}{|~}*{6}{|-}}
				 & GIST - channel 9  & - & \cellcolor[rgb]{  0.80,  0.18,  0.18}\textcolor{white}{\textbf{  0.03}*}  & - & - & \cellcolor[rgb]{  0.16,  0.16,  0.80}\textcolor{white}{\textbf{ -0.04}*}  \\ \hhline{*{1}{|~}*{6}{|-}}
				 & GIST - channel 10  & - & - & - & \cellcolor[rgb]{  0.16,  0.16,  0.80}\textcolor{white}{\textbf{ -0.03}}  & \cellcolor[rgb]{  0.16,  0.16,  0.80}\textcolor{white}{\textbf{ -0.02}}  \\ \hhline{*{1}{|~}*{6}{|-}}
				 & GIST - channel 11  & - & \cellcolor[rgb]{  0.80,  0.18,  0.18}\textcolor{white}{\textbf{  0.03}*}  & \cellcolor[rgb]{  0.80,  0.18,  0.18}\textcolor{white}{\textbf{  0.03}}  & - & \cellcolor[rgb]{  0.16,  0.16,  0.80}\textcolor{white}{\textbf{ -0.05}*}  \\ \hhline{*{1}{|~}*{6}{|-}}
				 & GIST - channel 12  & \cellcolor[rgb]{  0.80,  0.18,  0.18}\textcolor{white}{\textbf{  0.03}}  & - & - & - & \cellcolor[rgb]{  0.16,  0.16,  0.80}\textcolor{white}{\textbf{ -0.03}}  \\ \hhline{*{1}{|~}*{6}{|-}}
				 & GIST - channel 15  & - & - & \cellcolor[rgb]{  0.80,  0.18,  0.18}\textcolor{white}{\textbf{  0.04}*}  & - & \cellcolor[rgb]{  0.16,  0.16,  0.80}\textcolor{white}{\textbf{ -0.04}*}  \\ \hhline{*{1}{|~}*{6}{|-}}
				 & GIST - channel 21  & - & \cellcolor[rgb]{  0.80,  0.18,  0.18}\textcolor{white}{\textbf{  0.03}}  & - & - & \cellcolor[rgb]{  0.16,  0.16,  0.80}\textcolor{white}{\textbf{ -0.02}}  \\ \hhline{*{1}{|~}*{6}{|-}}
				 & GIST - channel 23  & - & \cellcolor[rgb]{  0.80,  0.18,  0.18}\textcolor{white}{\textbf{  0.02}}  & \cellcolor[rgb]{  0.80,  0.18,  0.18}\textcolor{white}{\textbf{  0.04}*}  & - & \cellcolor[rgb]{  0.16,  0.16,  0.80}\textcolor{white}{\textbf{ -0.03}*}  \\ \hline
				Faces
				 & number of faces  & \cellcolor[rgb]{  0.16,  0.16,  0.80}\textcolor{white}{\textbf{ -0.08}*}  & \cellcolor[rgb]{  0.80,  0.18,  0.18}\textcolor{white}{\textbf{  0.04}*}  & \cellcolor[rgb]{  0.80,  0.18,  0.18}\textcolor{white}{\textbf{  0.07}*}  & \cellcolor[rgb]{  0.80,  0.18,  0.18}\textcolor{white}{\textbf{  0.03}*}  & - \\ \hline
			\end{tabular}%
		}
	}
	\caption{Correlation scores (Spearman $\rho$ correlation coefficients) between features and personality trait scores. Red cells represent positive correlations while blue ones negative correlations; all filled cells are significant at $p<0.05$, cells with * are significant at $p<0.01$.}
	\label{table:corr}\vspace{-0.7cm}
\end{table*}

\begin{figure*}[!ht]
	\centering
	\begin{minipage}[b]{0.45\linewidth}
	\centering
		\includegraphics[width=0.7\linewidth]{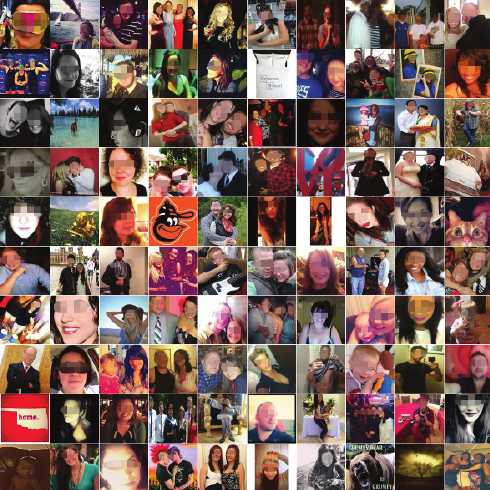}
		\caption{A random sampling of images of users with high level of Extraversion, blurred for privacy reasons.}
		\label{fig:Emax}
	\end{minipage}
	\quad
	\begin{minipage}[b]{0.45\linewidth}
	\centering
		\includegraphics[width=0.7\linewidth]{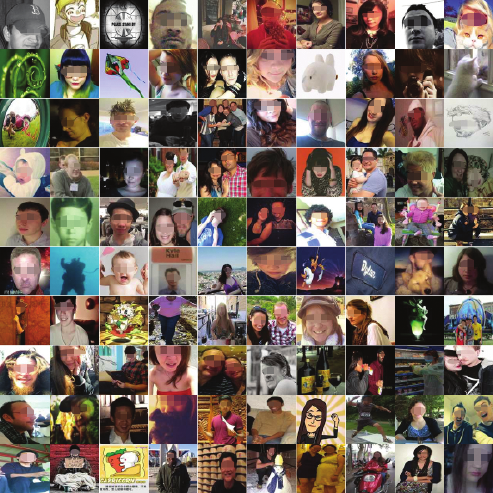}\vspace{-.2cm}
		\caption{A random sampling of images of users with low level of Extraversion, blurred for privacy reasons.}
		\label{fig:Emin}
	\end{minipage}\vspace{-.2cm}
\end{figure*}

Some considerations emerge and can be fully appreciated by looking at the table \textit{and then} checking the pictures of the users extracted randomly from the two extreme quartiles $Q_{1,3}$. Extroverts (see the Table~\ref{table:corr}, Fig.~\ref{fig:Emax} and~\ref{fig:Emin})% pictures have been blurred for privacy issues, in order to make the face unrecognizable) 
tend to exhibit many faces in their portraits, probably mirroring their inclination to socialize. This result confirm those found by~\cite{hall.al13,wu2014facebook}. The positive correlation with the \textit{rule of thirds} (which analyzes where the salient objects are located in the images, mimicking the rule adopted by professional photographers) states that face is located around the center of the picture, but not \emph{precisely in the center}. Positive associations are also present for warm colors: this may represent a shared stylistic inclination of extroverts, or simply it witness the fact the the faces in the images occupy a large region, making warm colors predominant. The positive correlation with \textit{wavelets, level of details} and the negative correlation with \textit{average region size, colorfulness} suggest that pictures present few sharp saturated regions, with uniform colors. 
In Fig.~\ref{fig:Emax} we report pictures of extroverts. Taking a look at these pictures we can observe that extroverts present a pleasant profile picture, with the face/faces not perfectly centered in the image. Although the images do not present rich variation of colors, we can notice the strong predominance of warm and bright colors. We can also notice the presence of few, but large, bright and sharp regions. Instead, in Fig.~\ref{fig:Emin} are shown pictures of introverts. We notice that images appear smooth and with dull colors (this last visible in the figure). The subject posts a picture where he/she is mostly alone, half bust, or with a drawing that can represent him/herself.

Neurotic people have a certain tendency of showing portraits with many colors (high colorfulness), with the presence of large regions, due to the positive association with texture features like \textit{GLCM}, aimed at the computation of the spatial relation between gray levels. Emotionally stable individuals tend to have sharp details in the picture, like natural scenes, due to the negative associations with \textit{GIST, wavelets, green} features. 

Agreeable users tend to have pictures with lots of faces, characterized by a strong level of warm color due to the presence of faces, and a positive association with \textit{rule of thirds}. The negative association with the \textit{average region size} suggests that unpleasant individuals tend to have pictures with large, homogeneous regions. Moreover, disagreeable ones tend to have pictures where they are portrayed alone. Further, it seems they do not choose carefully their pictures, as most of them are shot out or too bright because too prolonged exposure of the camera shutter.

Finally, conscientious individuals tend to have pictures with lots of faces, while open individuals tend to have pictures where they are alone.

%In Fig~\ref{fig:Amax}  we can observe that Agreeableness Facebook users are depict in portrait pictures or together with other people. Vice versa in Fig~\ref{fig:Amin} we report 

\subsection{Classification}\label{sec.classifica}
In the second study the goal is to predict the personality traits (that is, their quantizations with respect to the mean $\mu$ and the quartiles $Q_{1,3}$), exploiting one or more families of features. Therefore, we perform two separate binary classification tasks where the classes \textit{below the mean/above the mean} and \textit{below the first quartile/above the third quartile} are labeled with 0,1 respectively. It is worth noting that in the quartile task not all the images are considered both for the training and the testing, but only those in the specific quartile partition (about 6500 instances). 

As for the features, the idea is to consider those that correlate significantly with a particular trait, and use them jointly to guess the class; thus we perform a preliminary feature selection step reducing significantly the number of features (e.g. a subset of 248 CNN-based features is selected for the Extraversion classification task starting from the original set of 4096 CNN-based features). Specifically, we perform a correlation-based feature selection, where for each trait we retained only those features that resulted statistically significant. This approach is simple but
very effective and often used in the personality computing literature~\cite{vinciarelli2014survey, aran.gaticaperez13, biel.gaticaperez13}. 

As for the classifier, we use a logistic regressor%~\cite{freedman2009statistical}
, where the decision rule for minimizing the error is to predict $y^u=1$ if $P(y^u=1 |x_{test})>0.5$. In addition to use logistic regression, we tried other methods such as linear regression%~\cite{seber2012linear}
, LASSO%~\cite{tibshirani1996regression}
, stepwise regression%~\cite{draper1966applied}
, ridge regression obtaining worst results. 

All the classification experiments have been performed using an averaged Hold-Out protocol: the classifier is trained over the 75\% of the dataset and tested over the remaining 25\% of it, performing 10 repetitions with shuffled training and testing partitions, then computing accuracy and F1-measure for each iteration. 

We decided to keep a uniform distribution of the samples, thus we built a balanced training and testing set: given a trait, we found the largest class $C_1$, with cardinality $N$ ($M$ for the cardinality of the other class $C_2$); we select $N$ uniformly random samples from $C_1$, while we keep all the samples for $C_2$, obtaining two classes with same cardinality. We then proceed to split the dataset in the training and testing set: this procedure reduced the overall dataset to more than 7000 images for each trait. 

In Table~\ref{table:classres} we report the classification results with respect to the mean and quartile split with all the families of (correlating) features. We can first notice that Extraversion has the highest scores, and that the quartile split results are higher compared to the mean split, demonstrating that the scores near the mean value correspond to images less recognizable. We also performed a \textit{t-Student} test to confirm the significance of the classification results, testing that they are statistically above the chance level. 

In Table~\ref{table:classfamily} we report the classification results for each feature family, and for each combination of two and three feature families, with respect to the quartile binarization. The CNN-based features give the best classification results, the CA-based features are essentially equivalent to IATO-based features, and the PHOW-based ones exhibit the lowest performance.

The underlying reason for the success of the CNN-based features could be that of their large number with respect to the other families (e.g. 284 CNN-based features vs. 30 CA-based features, 50 IATO-based features and 172 PHOW-based features for Extraversion classification). By evaluating the combination of families of features, in Table~\ref{table:classfamily} one can observe that the combination CA + CNN + IATO give the best results together with PHOW + CNN + IATO and all the four families of features, thus witnessing the essential superiority of the pair CNN-IATO. 

\begin{table}[h]
	\centering
	\resizebox{1\linewidth}{!}{
\renewcommand{\arraystretch}{1.0}
{\normalsize
	\begin{tabular}{c|c|c|c|c|}
		\cline{2-5}
		& \multicolumn{2}{c|}{Mean Split} & \multicolumn{2}{c|}{Quartile Split} \\ \cline{2-5} 
		& Accuracy & F1 Measure & Accuracy & F1 Measure \\ \hline
		\multicolumn{1}{|c|}{O} & 0.55 & 0.56 & 0.60 & 0.60 \\ \hline
		\multicolumn{1}{|c|}{C} & 0.55 & 0.57 & 0.60 & 0.60 \\ \hline
		\multicolumn{1}{|c|}{E} & 0.56 & 0.56 & 0.62 & 0.62 \\ \hline
		\multicolumn{1}{|c|}{A} & 0.55 & 0.55 & 0.60 & 0.60 \\ \hline
		\multicolumn{1}{|c|}{N} & 0.55 & 0.55 & 0.60 & 0.60 \\ \hline
	\end{tabular}}}
	\caption{Classification scores with respect to mean split (left) and quartile split (right). \label{table:classres}}
\end{table}
\vspace{-0.3cm}

\begin{table}[!ht]
		\centering
		\resizebox{1\linewidth}{!}{
			\renewcommand{\arraystretch}{1.0}
	\begin{tabular}{l|c|c|c|c|c|}
		\cline{2-6}
		& \multicolumn{5}{c|}{\textbf{Mean Classification Accuracy}} \\ \cline{2-6} 
		\multicolumn{1}{c|}{} & \textbf{O} & \textbf{C} & \textbf{E} & \textbf{A} & \textbf{N} \\ \hline
		\multicolumn{6}{|c|}{\cellcolor[HTML]{C0C0C0}\textbf{One family of features}} \\ \hline
		\multicolumn{1}{|l|}{CA} & 0.55 & 0.52 & 0.55 & 0.53 & 0.53 \\ \hline
		\multicolumn{1}{|l|}{PHOW} & 0.54 & 0.54 & 0.54 & 0.54 & 0.54 \\ \hline
		\multicolumn{1}{|l|}{CNN} & \underline{\textbf{0.59}} & \underline{\textbf{0.60}} & \underline{\textbf{0.61}} & \underline{\textbf{0.60}} & \underline{\textbf{0.59}} \\ \hline
		\multicolumn{1}{|l|}{IATO} & 0.53 & 0.53 & 0.55 & 0.53 & 0.53 \\ \hline
		\multicolumn{6}{|c|}{\cellcolor[HTML]{C0C0C0}\textbf{Combination of two families of features}} \\ \hline
		\multicolumn{1}{|l|}{CA-PHOW} & 0.54 & 0.55 & 0.55 & 0.54 & 0.54 \\ \hline
		\multicolumn{1}{|l|}{CA-CNN} & 0.60 & 0.60 & 0.61 & 0.60 & 0.59 \\ \hline
		\multicolumn{1}{|l|}{CA-IATO} & 0.54 & 0.54 & 0.56 & 0.54 & 0.54 \\ \hline
		\multicolumn{1}{|l|}{PHOW-CNN} & 0.59 & 0.60 & 0.62 & 0.61 & 0.60 \\ \hline
		\multicolumn{1}{|l|}{PHOW-IATO} & 0.54 & 0.54 & 0.56 & 0.54 & 0.55 \\ \hline
		\multicolumn{1}{|l|}{CNN-IATO} & \underline{\textbf{0.60}} & \underline{\textbf{0.60}} & \underline{\textbf{0.62}} & \underline{\textbf{0.60}} & \underline{\textbf{0.60}} \\ \hline
		\multicolumn{6}{|c|}{\cellcolor[HTML]{C0C0C0}\textbf{Combination of three families of features}} \\ \hline
		\multicolumn{1}{|l|}{CA-PHOW-CNN} & 0.60 & 0.60 & 0.61 & 0.60 & 0.59 \\ \hline
		\multicolumn{1}{|l|}{CA-PHOW-IATO} & 0.55 & 0.55 & 0.55 & 0.55 & 0.55 \\ \hline
		\multicolumn{1}{|l|}{CA-CNN-IATO} & \underline{\textbf{0.61}} & 0.60 & \underline{\textbf{0.62}} & \underline{\textbf{0.60}} & \underline{\textbf{0.60}} \\ \hline
		\multicolumn{1}{|l|}{PHOW-CNN-IATO} & 0.60 & \underline{\textbf{0.61}} & \underline{\textbf{0.62}} & \underline{\textbf{0.60}} &\underline{ \textbf{0.60}} \\ \hline
		\multicolumn{6}{|c|}{\cellcolor[HTML]{C0C0C0}\textbf{All families of features}} \\ \hline
		\multicolumn{1}{|l|}{CA-PHOW-CNN-IATO} & \underline{\textbf{0.60}} & \underline{\textbf{0.60}} & \underline{\textbf{0.62}} & \underline{\textbf{0.60}} & \underline{\textbf{0.60}} \\ \hline
	\end{tabular}
}
	\caption{Mean classification accuracy with respect to quartile split for each family of feature and combinations of families of features.\label{table:classfamily}\vspace{-0.8cm}}
\end{table}		
		
\subsection{Comparing computer-based and human-based classification}
In this last experiment, we compare the accuracy of human and computer-based personality classification from Facebook profile pictures. As first step, we build a ``reduced'' test dataset, focusing on Extraversion and Neuroticism traits (since they get higher scores as for the correlational analysis and the classification task) and by sampling 150 subjects (75 from the first quartile and 75 from the third quartile of the two traits). 
The 150 Facebook users' images have been organized on a web interface, where a rater can select a score between 1 (introvert/emotionally stable) and 5 (extrovert/neurotic). 23 participants have been asked to fill the score for each of the 150 Facebook users, while observing his/her Facebook profile picture. We computed the agreement among the raters with Krippendorf's $\alpha$~\cite{kripp04}, a reliability coefficient suitable for a wide variety of assessments and robust to small sample sizes. The value of $\alpha$ is 0.34 for Extraversion and 0.26 for Neuroticism. The values are statistically significant. After that, the selected scores of the 150 Facebook users by each rater have been used as they were provided by a classifier, computing the mean and maximum accuracy and F1-measure when predicting the Extraversion or Neuroticism trait, as reported in Table~\ref{table:userstudy}. 
Second, we used those 150 users of the interface as testing set of our previous computer-based approach (see Sec.~\ref{sec.classifica}), and the remaining users of the trait under examination as the training set. We computed the accuracy and F1-measure of the classification and reported them in Table~\ref{table:userstudy}. 
We notice that the classification performed by the \textit{computer} significantly outperforms the average \textit{human-based} classification both for Extraversion and Neuroticism. Thus, our results support those recently obtained by Youyou \textit{et al.}~\cite{youyou.al15}, wherein automatic predictions based on Facebook Likes are more accurate than those made by the participants' Facebook friends.
%DAVID This study isn't quite the same as our Youyou one, because we used friend-ratings of personality as the comparison. So they were people that knew them in real life, and were able to use whatever 'data' they wanted to make a decision. In this study, raters do not know the rated, so the profile pic is all they can use.

\begin{table}[h]
	\centering 
	\resizebox{1\linewidth}{!}{ 
		\renewcommand{\arraystretch}{1.0} 
		{\normalsize 
	\begin{tabular}{c|c|c|c|c|}
		\cline{2-5}
			& \multicolumn{2}{c|}{User study Quartile} & \multicolumn{2}{c|}{User Study Quartile } \\
			& \multicolumn{2}{c|}{HUMAN} & \multicolumn{2}{c|}{ MACHINE}\\ \cline{2-5} 
			& Accuracy & F1 Measure & Mean Accuracy & Mean F1-Measure \\ \hline
			\multicolumn{1}{|c|}{E} & 0.60 & 0.57 & 0.68 & 0.72 \\ \hline
			\multicolumn{1}{|c|}{N} & 0.58 & 0.60 & 0.69 & 0.67 \\ \hline
		\end{tabular}}}
		\caption{Comparison between classification performed by \textit{human} and \textit{machine} on a reduced dataset of 150 pictures/users.\label{table:userstudy}\vspace{-0.7cm}}
\end{table}

%Considering the fact that we used a dataset similar in size to the one used by~\cite{celli2014automatic}, we outperformed their results on Extraversion (F1 = 0.615) and Neuroticism (F1 = 0.609), and we noticed a better performance of our classifiers on such a small data size. We interpret this fact as the modelization of personality is easier on few pictures and harder on a large amount of data. \\
%DAVID I don't understand where this statement comes from. It reads to me like you're saying:
%1. we had the same amount of data as them
%2. our performance was better
%3. so modeling personality is easier when the dataset is smaller
According to previous works in social psychology~\cite{funder2012}, the accuracy of \textit{human} personality judgements depends on the available amount of relevant information and on the ability of the person to detect and use it in a correct way. Under this perspective, computers have several major advantages over humans. First of all, computers have the capacity of storing a tremendous amount of information, which is difficult for humans to access. Then, computers use these information in order to optimize the judgmental accuracy, while humans are often affected by various motivational biases~\cite{vazire2011}.

\section{Discussion}\label{disc}
In the current paper we explore how self-assessed personality traits can be automatically inferred by just looking at the Facebook profile pictures. We compare four different visual features' families in the classification of the Big Five personality traits, and several interesting insights have emerged, which pave the way for future work.
Among the different families of features taken into account, those that correlate the most come from the Computational Aesthetics (CA) field; these cues are highly interpretable and model medium/high-level patterns in the image such as the number of occurrences of a particular color name or the textural properties. However, it is worth noting that the absolute value of the correlations is low, meaning that more can be done, for example crafting features specifically suited for capturing personality facets. In this sense, deep learning appears to be a valid answer, provided that the number of images to learn is much bigger. Indeed, when it comes to the classification task, CNN-based features obtain the best performance; presumably, this is due to their large number with respect to other feature families (i.e. two orders of magnitude larger than CA-based features).

Interestingly, dense/global features, together with faces, play a crucial role in the personality classification task. This suggests that the context of a picture is also important for determining the personality trait of an individual: as a consequence, future studies can focus on features which characterize the face of a subject such as expression, gaze direction, head pose estimation  and emotions \cite{tettegah2016emotions}, but also the scene in which he/she is immersed.

Finally, a comparison between humans and algorithms in the detection of personality traits from profile pictures revealed that the algorithms perform better than humans, a finding that is supported by previous recent literature~\cite{youyou.al15}. This result definitely encourages a larger and systematic user study, aimed at distilling those aspects that contribute in performing a better guess and those that work as confounding elements, both for the human and the machine.

Our work has also some limitations: first of all, we take into account only one profile picture per user. In the future it would be interesting to extend the dataset using a larger number of profile pictures for each user, in order to investigate if the patterns are stable and consistent over time or whether the frequency of profile picture's change could be helpful in the classification task. Another limitation is that we try to build a general model from all the profile pictures in the dataset. There is a large variety of subjects depicted in the profile pictures, including symbols and animals. In the future it would be interesting to test models extracted from more homogeneous datasets, for example selecting only profile pictures containing faces. 

Further, we based most of our results regarding correlations and classification on the hand-crafted features as we retained that, other than having higher correlations, they are more interpretable from a human and psychological point of view. However we saw that when using deep learning features, we achieve the best results in classification. In the future we are interested in exploiting more deep learning approaches, with the aim of building a deep model for each personality trait, able to better generalize and disentangle the factors characterizing each trait and that links each of them to the profile images of the users. With this purpose, however, it will be necessary to build and extrapolate interpretable visualizations and concepts from the extracted features that most of the time are used as black box and passed as input to classifiers. In comparison with this work it would be also interesting to explore whether using deep learning approaches we will achieve the same results for what concern the concepts that shape each personality trait.

Finally, we do not investigate whether personality types can be predicted also from the banner picture or from pictures posted on the individual's wall. According to recent literature in social psychology \cite{leary2011personality}, profile pictures seem to be the most important part of the self presentation in social networks. However, the information from other types of Facebook pictures may be helpful in the classification. Finally, the usage of external assessments of personality traits can be also valuable~\cite{segalin2016AffComp}, namely in the sense of calculating the correlations and the classification scores with respect to the label given by other people to the profile of an individual. We leave these open issues for future work.

Overall, an important message remarked by our paper is that unconsciously we are sharing on social networks more information than what we think; an apparently simple action, like uploading a profile picture on Facebook or other social networks, may unveil some aspects of our personality and this in turns may activate several possible business applications.

A first example is in the field of marketing, where advertising and recommendation activities and systems can be customized based on the information retrieved from our Facebook profile picture. Another possible application is in the field of social media monitoring, where personality can play a role in understanding people's intentions beside the usage of sentiment analysis techniques. Again, other potential fields include (i) people's well-being, where personality classification could be exploited to help early detection of psychopathology; (ii) human resources, where classification of personality types could help monitoring and selecting job candidates on a large scale; (iii) dating, where personality could be exploited for profile matching; and (iv) the field of financial technologies, where personality types could be useful for computing more accurate personal credit scoring metrics.

\section{Conclusions}\label{conc}
In this work we investigate automatic personality classification from Facebook profile pictures. Specifically, we analyze the effectiveness of four families of visual features (Computational Aesthetics, Pyramid Histogram Of visual Words, Image Analysis TOol, and Convolutional Neural Networks based features), and we discuss some human interpretable patterns that explain the personality traits of the individuals. For example, agreeable individuals and extroverts tend to have warm colored pictures and to exhibit many faces in their portraits, mirroring
their inclination to socialize; while neurotic people have a prevalence of pictures of indoor places. Moreover, we propose a classification approach of personality traits from these visual features. Finally, we focus on two personality traits, Extraversion and Neuroticism, and we compare the performance of our classification approach to the one obtained by human raters. The results show that computer-based classifications are significantly more accurate than averaged human-based classifications for these two traits.

Although there is still room for improvement in the classification of personality traits from profile pictures, our results show that Facebook profile pictures convey relevant information for classifying the personality traits of an individual.

\section{Acknowledgments}
The work of Fabio Celli and Bruno Lepri was partly supported by the EIT High Impact Initiative Street Smart Retail (HII SSR).

\bibliographystyle{ACM-Reference-Format}
\bibliography{bruno,sigproc,submission}  % sigproc.bib is the name of the Bibliography in this case


\begin{thebibliography}{00}

%%% the Web address.

\ifx \showCODEN    \undefined \def \showCODEN     #1{\unskip}     \fi
\ifx \showDOI      \undefined \def \showDOI       #1{#1}\fi
\ifx \showISBNx    \undefined \def \showISBNx     #1{\unskip}     \fi
\ifx \showISBNxiii \undefined \def \showISBNxiii  #1{\unskip}     \fi
\ifx \showISSN     \undefined \def \showISSN      #1{\unskip}     \fi
\ifx \showLCCN     \undefined \def \showLCCN      #1{\unskip}     \fi
\ifx \shownote     \undefined \def \shownote      #1{#1}          \fi
\ifx \showarticletitle \undefined \def \showarticletitle #1{#1}   \fi
\ifx \showURL      \undefined \def \showURL       {\relax}        \fi
% invisible to TeX
\providecommand\bibfield[2]{#2}
\providecommand\bibinfo[2]{#2}
\providecommand\natexlab[1]{#1}
\providecommand\showeprint[2][]{arXiv:#2}

\bibitem[\protect\citeauthoryear{Aran and Gatica-Perez}{Aran and
  Gatica-Perez}{2013}]%
        {aran.gaticaperez13}
\bibfield{author}{\bibinfo{person}{Oya Aran} {and} \bibinfo{person}{Daniel
  Gatica-Perez}.} \bibinfo{year}{2013}\natexlab{}.
\newblock \showarticletitle{Cross-domain personality prediction: from video
  blogs to small group meetings}. In \bibinfo{booktitle}{{\em Proceedings of
  ICMI}}. ACM, \bibinfo{pages}{127--130}.
\newblock


\bibitem[\protect\citeauthoryear{Batrinca, Mana, Lepri, Pianesi, and
  Sebe}{Batrinca et~al\mbox{.}}{2011}]%
        {batrinca2011please}
\bibfield{author}{\bibinfo{person}{Ligia~Maria Batrinca},
  \bibinfo{person}{Nadia Mana}, \bibinfo{person}{Bruno Lepri},
  \bibinfo{person}{Fabio Pianesi}, {and} \bibinfo{person}{Nicu Sebe}.}
  \bibinfo{year}{2011}\natexlab{}.
\newblock \showarticletitle{Please, tell me about yourself: automatic
  personality assessment using short self-presentations}. In
  \bibinfo{booktitle}{{\em Proceedings of the 13th international conference on
  multimodal interfaces}}. ACM, \bibinfo{pages}{255--262}.
\newblock


\bibitem[\protect\citeauthoryear{Biel and Gatica-Perez}{Biel and
  Gatica-Perez}{2013}]%
        {biel.gaticaperez13}
\bibfield{author}{\bibinfo{person}{J Biel} {and} \bibinfo{person}{Daniel
  Gatica-Perez}.} \bibinfo{year}{2013}\natexlab{}.
\newblock \showarticletitle{The youtube lens: Crowdsourced personality
  impressions and audiovisual analysis of vlogs}.
\newblock \bibinfo{journal}{{\em Multimedia, IEEE Trans. on\/}}
  \bibinfo{volume}{15}, \bibinfo{number}{1} (\bibinfo{year}{2013}),
  \bibinfo{pages}{41--55}.
\newblock


\bibitem[\protect\citeauthoryear{Bosch, Zisserman, and Muoz}{Bosch
  et~al\mbox{.}}{2007}]%
        {bosch2007image}
\bibfield{author}{\bibinfo{person}{Anna Bosch}, \bibinfo{person}{Andrew
  Zisserman}, {and} \bibinfo{person}{Xavier Muoz}.}
  \bibinfo{year}{2007}\natexlab{}.
\newblock \showarticletitle{Image classification using random forests and
  ferns}. In \bibinfo{booktitle}{{\em Proceedings of IEEE ICCV}}.
  \bibinfo{pages}{1--8}.
\newblock


\bibitem[\protect\citeauthoryear{Celli, Bruni, and Lepri}{Celli
  et~al\mbox{.}}{2014}]%
        {celli2014automatic}
\bibfield{author}{\bibinfo{person}{Fabio Celli}, \bibinfo{person}{Elia Bruni},
  {and} \bibinfo{person}{Bruno Lepri}.} \bibinfo{year}{2014}\natexlab{}.
\newblock \showarticletitle{Automatic Personality and Interaction Style
  Recognition from Facebook Profile Pictures}. In \bibinfo{booktitle}{{\em
  Proceedings of ACM-MM}}. \bibinfo{pages}{1101--1104}.
\newblock


\bibitem[\protect\citeauthoryear{Chittaranjan, Blom, and
  Gatica-Perez}{Chittaranjan et~al\mbox{.}}{2013}]%
        {chittaranjan.al13}
\bibfield{author}{\bibinfo{person}{Gokul Chittaranjan}, \bibinfo{person}{Jan
  Blom}, {and} \bibinfo{person}{Daniel Gatica-Perez}.}
  \bibinfo{year}{2013}\natexlab{}.
\newblock \showarticletitle{Mining large-scale smartphone data for personality
  studies}.
\newblock \bibinfo{journal}{{\em Personal and Ubiquitous Computing\/}}
  \bibinfo{volume}{17}, \bibinfo{number}{3} (\bibinfo{year}{2013}),
  \bibinfo{pages}{433--450}.
\newblock


\bibitem[\protect\citeauthoryear{Costa and McCrae}{Costa and McCrae}{1985}]%
        {costa1985neo}
\bibfield{author}{\bibinfo{person}{P.T. Costa} {and} \bibinfo{person}{R.R.
  McCrae}.} \bibinfo{year}{1985}\natexlab{}.
\newblock \bibinfo{booktitle}{{\em The NEO Personality Inventory: Manual Form S
  and Form R}}.
\newblock \bibinfo{publisher}{Psychological Assessment Resources}.
\newblock
\showURL{%
\url{http://books.google.it/books?id=q6oFHAAACAAJ}}


\bibitem[\protect\citeauthoryear{Cristani, Vinciarelli, Segalin, and
  Perina}{Cristani et~al\mbox{.}}{2013}]%
        {Cristani:Flickr:ACMMM:2013}
\bibfield{author}{\bibinfo{person}{Marco Cristani}, \bibinfo{person}{Alessandro
  Vinciarelli}, \bibinfo{person}{Cristina Segalin}, {and}
  \bibinfo{person}{Alessandro Perina}.} \bibinfo{year}{2013}\natexlab{}.
\newblock \showarticletitle{Unveiling the Multimedia Unconscious: Implicit
  Cognitive Processes and Multimedia Content Analysis}. In
  \bibinfo{booktitle}{{\em Proceedings of ACM-MM}}. \bibinfo{pages}{213--222}.
\newblock
\newblock
\shownote{Brand New Idea paper.}


\bibitem[\protect\citeauthoryear{Csurka, Dance, Fan, Willamowski, and
  Bray}{Csurka et~al\mbox{.}}{2004}]%
        {csurka2004visual}
\bibfield{author}{\bibinfo{person}{Gabriella Csurka},
  \bibinfo{person}{Christopher Dance}, \bibinfo{person}{Lixin Fan},
  \bibinfo{person}{Jutta Willamowski}, {and} \bibinfo{person}{C{\'e}dric
  Bray}.} \bibinfo{year}{2004}\natexlab{}.
\newblock \showarticletitle{Visual categorization with bags of keypoints}. In
  \bibinfo{booktitle}{{\em Workshop on Statistical Learning in Computer Vision,
  ECCV}}, Vol.~\bibinfo{volume}{1}. Prague, \bibinfo{pages}{1--2}.
\newblock


\bibitem[\protect\citeauthoryear{Datta, Joshi, Li, and Wang}{Datta
  et~al\mbox{.}}{2006}]%
        {Datta06}
\bibfield{author}{\bibinfo{person}{R. Datta}, \bibinfo{person}{D. Joshi},
  \bibinfo{person}{J. Li}, {and} \bibinfo{person}{J. Wang}.}
  \bibinfo{year}{2006}\natexlab{}.
\newblock \showarticletitle{Studying Aesthetics in Photographic Images Using a
  Computational Approach}.
\newblock In \bibinfo{booktitle}{{\em Proceedings of ECCV}}.
  Vol.~\bibinfo{volume}{3953}. \bibinfo{publisher}{Springer Verlag},
  \bibinfo{pages}{288--301}.
\newblock


\bibitem[\protect\citeauthoryear{Deng, Dong, Socher, Li, Li, and Fei-Fei}{Deng
  et~al\mbox{.}}{2009}]%
        {deng2009imagenet}
\bibfield{author}{\bibinfo{person}{Jia Deng}, \bibinfo{person}{Wei Dong},
  \bibinfo{person}{Richard Socher}, \bibinfo{person}{Li-Jia Li},
  \bibinfo{person}{Kai Li}, {and} \bibinfo{person}{Li Fei-Fei}.}
  \bibinfo{year}{2009}\natexlab{}.
\newblock \showarticletitle{Imagenet: A large-scale hierarchical image
  database}. In \bibinfo{booktitle}{{\em Proceedings of IEEE CVPR}}.
  \bibinfo{pages}{248--255}.
\newblock


\bibitem[\protect\citeauthoryear{Diefenbach and Christoforakos}{Diefenbach and
  Christoforakos}{2017}]%
        {diefenbach2017selfie}
\bibfield{author}{\bibinfo{person}{Sarah Diefenbach} {and}
  \bibinfo{person}{Lara Christoforakos}.} \bibinfo{year}{2017}\natexlab{}.
\newblock \showarticletitle{The Selfie Paradox: Nobody Seems to Like Them Yet
  Everyone Has Reasons to Take Them. An Exploration of Psychological Functions
  of Selfies in Self-Presentation}.
\newblock \bibinfo{journal}{{\em Frontiers in psychology\/}}
  \bibinfo{volume}{8} (\bibinfo{year}{2017}).
\newblock


\bibitem[\protect\citeauthoryear{Evans, Gosling, and Carroll}{Evans
  et~al\mbox{.}}{2008}]%
        {EvansGoslingCarroll2008}
\bibfield{author}{\bibinfo{person}{D.C. Evans}, \bibinfo{person}{S.D. Gosling},
  {and} \bibinfo{person}{A. Carroll}.} \bibinfo{year}{2008}\natexlab{}.
\newblock \showarticletitle{What elements of an online social networking
  profile predict target-rater agreement in personality impressions}. In
  \bibinfo{booktitle}{{\em Proceedings of AAAI ICWSM}}.
  \bibinfo{pages}{45--50}.
\newblock


\bibitem[\protect\citeauthoryear{Ferwerda, Schedl, and Tkalcic}{Ferwerda
  et~al\mbox{.}}{2016}]%
        {ferwerda2016}
\bibfield{author}{\bibinfo{person}{B. Ferwerda}, \bibinfo{person}{M. Schedl},
  {and} \bibinfo{person}{M. Tkalcic}.} \bibinfo{year}{2016}\natexlab{}.
\newblock \showarticletitle{Using Instagram picture features to predict users'
  personality}.
\newblock In \bibinfo{booktitle}{{\em MultiMedia Modeling}},
  \bibfield{editor}{\bibinfo{person}{Q.~Tian}, \bibinfo{person}{N.~Sebe},
  \bibinfo{person}{G.-J. Qi}, \bibinfo{person}{B.~Huet},
  \bibinfo{person}{R.~Hong}, {and} \bibinfo{person}{X.~Liu}} (Eds.).
  \bibinfo{publisher}{Springer International Publishing},
  \bibinfo{pages}{850--861}.
\newblock


\bibitem[\protect\citeauthoryear{Funder}{Funder}{2012}]%
        {funder2012}
\bibfield{author}{\bibinfo{person}{D.C. Funder}.}
  \bibinfo{year}{2012}\natexlab{}.
\newblock \showarticletitle{Accurate personality judgment}.
\newblock \bibinfo{journal}{{\em Current Directions in Psychological
  Science\/}} \bibinfo{volume}{21}, \bibinfo{number}{3} (\bibinfo{year}{2012}),
  \bibinfo{pages}{1--18}.
\newblock


\bibitem[\protect\citeauthoryear{Georgescu}{Georgescu}{2002}]%
        {Georgescu02}
\bibfield{author}{\bibinfo{person}{C.M. Georgescu}.}
  \bibinfo{year}{2002}\natexlab{}.
\newblock \showarticletitle{Synergism in low level vision}. In
  \bibinfo{booktitle}{{\em Proceedings of ICPR}}. \bibinfo{pages}{150--155}.
\newblock


\bibitem[\protect\citeauthoryear{Goffman}{Goffman}{1959}]%
        {goffman1959}
\bibfield{author}{\bibinfo{person}{E. Goffman}.}
  \bibinfo{year}{1959}\natexlab{}.
\newblock \bibinfo{booktitle}{{\em The presentation of self in everyday life}}.
\newblock \bibinfo{publisher}{Garden City, NY Double Day}.
\newblock


\bibitem[\protect\citeauthoryear{Golbeck, Robles, Edmondson, and
  Turner}{Golbeck et~al\mbox{.}}{2011b}]%
        {golbeck.al11b}
\bibfield{author}{\bibinfo{person}{Jennifer Golbeck}, \bibinfo{person}{Cristina
  Robles}, \bibinfo{person}{Michon Edmondson}, {and} \bibinfo{person}{Karen
  Turner}.} \bibinfo{year}{2011}\natexlab{b}.
\newblock \showarticletitle{Predicting personality from Twitter}. In
  \bibinfo{booktitle}{{\em Proceedings of IEEE SocialCom}}.
  \bibinfo{pages}{149--156}.
\newblock


\bibitem[\protect\citeauthoryear{Golbeck, Robles, and Turner}{Golbeck
  et~al\mbox{.}}{2011a}]%
        {golbeck2011predicting}
\bibfield{author}{\bibinfo{person}{Jennifer Golbeck}, \bibinfo{person}{Cristina
  Robles}, {and} \bibinfo{person}{Karen Turner}.}
  \bibinfo{year}{2011}\natexlab{a}.
\newblock \showarticletitle{Predicting personality with social media}. In
  \bibinfo{booktitle}{{\em CHI'11 extended abstracts on human factors in
  computing systems}}. ACM, \bibinfo{pages}{253--262}.
\newblock


\bibitem[\protect\citeauthoryear{Goldberg, Johnson, Eber, Hogan, Ashton,
  Cloninger, and Gough}{Goldberg et~al\mbox{.}}{2006}]%
        {goldberg2006international}
\bibfield{author}{\bibinfo{person}{Lewis~R Goldberg}, \bibinfo{person}{John~A
  Johnson}, \bibinfo{person}{Herbert~W Eber}, \bibinfo{person}{Robert Hogan},
  \bibinfo{person}{Michael~C Ashton}, \bibinfo{person}{C~Robert Cloninger},
  {and} \bibinfo{person}{Harrison~G Gough}.} \bibinfo{year}{2006}\natexlab{}.
\newblock \showarticletitle{The international personality item pool and the
  future of public-domain personality measures}.
\newblock \bibinfo{journal}{{\em Journal of Research in Personality\/}}
  \bibinfo{volume}{40}, \bibinfo{number}{1} (\bibinfo{year}{2006}),
  \bibinfo{pages}{84--96}.
\newblock


\bibitem[\protect\citeauthoryear{Gonzales and Hancock}{Gonzales and
  Hancock}{2008}]%
        {GonzalesHancock2008}
\bibfield{author}{\bibinfo{person}{A.L. Gonzales} {and} \bibinfo{person}{J.T.
  Hancock}.} \bibinfo{year}{2008}\natexlab{}.
\newblock \showarticletitle{Identity shift in computer-mediated environments}.
\newblock \bibinfo{journal}{{\em Media Psychology\/}} \bibinfo{volume}{11},
  \bibinfo{number}{2} (\bibinfo{year}{2008}), \bibinfo{pages}{167--185}.
\newblock


\bibitem[\protect\citeauthoryear{Gosling, Gaddis, and Vazire}{Gosling
  et~al\mbox{.}}{2007}]%
        {GoslingGaddisVazire2007}
\bibfield{author}{\bibinfo{person}{S.D. Gosling}, \bibinfo{person}{S. Gaddis},
  {and} \bibinfo{person}{S. Vazire}.} \bibinfo{year}{2007}\natexlab{}.
\newblock \showarticletitle{Personality impressions based on Facebook
  profiles}. In \bibinfo{booktitle}{{\em Proceedings of AAAI ICWSM}},
  Vol.~\bibinfo{volume}{7}. \bibinfo{pages}{1--4}.
\newblock


\bibitem[\protect\citeauthoryear{Hall and Pennington}{Hall and
  Pennington}{2013}]%
        {hall2013self}
\bibfield{author}{\bibinfo{person}{Jeffrey~A Hall} {and}
  \bibinfo{person}{Natalie Pennington}.} \bibinfo{year}{2013}\natexlab{}.
\newblock \showarticletitle{Self-monitoring, honesty, and cue use on Facebook:
  The relationship with user extraversion and conscientiousness}.
\newblock \bibinfo{journal}{{\em Computers in Human Behavior\/}}
  \bibinfo{volume}{29}, \bibinfo{number}{4} (\bibinfo{year}{2013}),
  \bibinfo{pages}{1556--1564}.
\newblock


\bibitem[\protect\citeauthoryear{Hall, Pennington, and Lueders}{Hall
  et~al\mbox{.}}{2013}]%
        {hall.al13}
\bibfield{author}{\bibinfo{person}{Jeffrey~A Hall}, \bibinfo{person}{Natalie
  Pennington}, {and} \bibinfo{person}{Allyn Lueders}.}
  \bibinfo{year}{2013}\natexlab{}.
\newblock \showarticletitle{Impression management and formation on Facebook: A
  lens model approach}.
\newblock \bibinfo{journal}{{\em New Media \& Society\/}}
  (\bibinfo{year}{2013}), \bibinfo{pages}{1--25}.
\newblock


\bibitem[\protect\citeauthoryear{Jayagopi, Hung, Yeo, and
  Gatica-Perez}{Jayagopi et~al\mbox{.}}{2009}]%
        {jayagopi2009modeling}
\bibfield{author}{\bibinfo{person}{Dinesh~Babu Jayagopi},
  \bibinfo{person}{Hayley Hung}, \bibinfo{person}{Chuohao Yeo}, {and}
  \bibinfo{person}{Daniel Gatica-Perez}.} \bibinfo{year}{2009}\natexlab{}.
\newblock \showarticletitle{Modeling dominance in group conversations using
  nonverbal activity cues}.
\newblock \bibinfo{journal}{{\em IEEE Transactions on Audio, Speech, and
  Language Processing\/}} \bibinfo{volume}{17}, \bibinfo{number}{3}
  (\bibinfo{year}{2009}), \bibinfo{pages}{501--513}.
\newblock


\bibitem[\protect\citeauthoryear{Jia, Shelhamer, Donahue, Karayev, Long,
  Girshick, Guadarrama, and Darrell}{Jia et~al\mbox{.}}{2014}]%
        {jia2014caffe}
\bibfield{author}{\bibinfo{person}{Yangqing Jia}, \bibinfo{person}{Evan
  Shelhamer}, \bibinfo{person}{Jeff Donahue}, \bibinfo{person}{Sergey Karayev},
  \bibinfo{person}{Jonathan Long}, \bibinfo{person}{Ross Girshick},
  \bibinfo{person}{Sergio Guadarrama}, {and} \bibinfo{person}{Trevor Darrell}.}
  \bibinfo{year}{2014}\natexlab{}.
\newblock \showarticletitle{Caffe: Convolutional Architecture for Fast Feature
  Embedding}.
\newblock \bibinfo{journal}{{\em arXiv preprint arXiv:1408.5093\/}}
  (\bibinfo{year}{2014}).
\newblock


\bibitem[\protect\citeauthoryear{Joshi, Gunes, and Goecke}{Joshi
  et~al\mbox{.}}{2014}]%
        {joshi2014automatic}
\bibfield{author}{\bibinfo{person}{Jyoti Joshi}, \bibinfo{person}{Hatice
  Gunes}, {and} \bibinfo{person}{Roland Goecke}.}
  \bibinfo{year}{2014}\natexlab{}.
\newblock \showarticletitle{Automatic prediction of perceived traits using
  visual cues under varied situational context}. In \bibinfo{booktitle}{{\em
  Proceedings of IEEE ICPR}}. \bibinfo{pages}{2855--2860}.
\newblock


\bibitem[\protect\citeauthoryear{Kosinski, Bachrach, Kohli, Stillwell, and
  Graepel}{Kosinski et~al\mbox{.}}{2013a}]%
        {kosinski.al13}
\bibfield{author}{\bibinfo{person}{Michal Kosinski}, \bibinfo{person}{Yoram
  Bachrach}, \bibinfo{person}{Pushmeet Kohli}, \bibinfo{person}{David
  Stillwell}, {and} \bibinfo{person}{Thore Graepel}.}
  \bibinfo{year}{2013}\natexlab{a}.
\newblock \showarticletitle{Manifestations of user personality in website
  choice and behaviour on online social networks}.
\newblock \bibinfo{journal}{{\em Machine Learning\/}} (\bibinfo{year}{2013}),
  \bibinfo{pages}{1--24}.
\newblock


\bibitem[\protect\citeauthoryear{Kosinski, Stillwell, and Graepel}{Kosinski
  et~al\mbox{.}}{2013b}]%
        {kosinski2013private}
\bibfield{author}{\bibinfo{person}{Michal Kosinski}, \bibinfo{person}{David
  Stillwell}, {and} \bibinfo{person}{Thore Graepel}.}
  \bibinfo{year}{2013}\natexlab{b}.
\newblock \showarticletitle{Private traits and attributes are predictable from
  digital records of human behavior}.
\newblock \bibinfo{journal}{{\em PNAS\/}} \bibinfo{volume}{110},
  \bibinfo{number}{15} (\bibinfo{year}{2013}), \bibinfo{pages}{5802--5805}.
\newblock


\bibitem[\protect\citeauthoryear{Krippendorff}{Krippendorff}{2004}]%
        {kripp04}
\bibfield{author}{\bibinfo{person}{K. Krippendorff}.}
  \bibinfo{year}{2004}\natexlab{}.
\newblock \showarticletitle{Reliability in Content Analysis}.
\newblock \bibinfo{journal}{{\em Human Communication Research\/}}
  \bibinfo{volume}{30}, \bibinfo{number}{3} (\bibinfo{year}{2004}),
  \bibinfo{pages}{411--433}.
\newblock


\bibitem[\protect\citeauthoryear{Krizhevsky, Sutskever, and Hinton}{Krizhevsky
  et~al\mbox{.}}{2012}]%
        {krizhevsky2012imagenet}
\bibfield{author}{\bibinfo{person}{Alex Krizhevsky}, \bibinfo{person}{Ilya
  Sutskever}, {and} \bibinfo{person}{Geoffrey~E Hinton}.}
  \bibinfo{year}{2012}\natexlab{}.
\newblock \showarticletitle{Imagenet classification with deep convolutional
  neural networks}. In \bibinfo{booktitle}{{\em NIPS}}.
  \bibinfo{pages}{1097--1105}.
\newblock


\bibitem[\protect\citeauthoryear{Leary and Allen}{Leary and Allen}{2011}]%
        {leary2011personality}
\bibfield{author}{\bibinfo{person}{Mark~R Leary} {and}
  \bibinfo{person}{Ashley~Batts Allen}.} \bibinfo{year}{2011}\natexlab{}.
\newblock \showarticletitle{Personality and Persona: Personality Processes in
  Self-Presentation}.
\newblock \bibinfo{journal}{{\em Journal of personality\/}}
  \bibinfo{volume}{79}, \bibinfo{number}{6} (\bibinfo{year}{2011}),
  \bibinfo{pages}{1191--1218}.
\newblock


\bibitem[\protect\citeauthoryear{LeCun and Bengio}{LeCun and Bengio}{1995}]%
        {lecun1995convolutional}
\bibfield{author}{\bibinfo{person}{Yann LeCun} {and} \bibinfo{person}{Yoshua
  Bengio}.} \bibinfo{year}{1995}\natexlab{}.
\newblock \showarticletitle{Convolutional networks for images, speech, and time
  series}.
\newblock \bibinfo{journal}{{\em The Handbook of Brain Theory and Neural
  Networks\/}}  \bibinfo{volume}{3361} (\bibinfo{year}{1995}),
  \bibinfo{pages}{310}.
\newblock


\bibitem[\protect\citeauthoryear{Lepri, Subramanian, Kalimeri, Staiano,
  Pianesi, and Sebe}{Lepri et~al\mbox{.}}{2012}]%
        {lepri.al12}
\bibfield{author}{\bibinfo{person}{Bruno Lepri}, \bibinfo{person}{Ramanathan
  Subramanian}, \bibinfo{person}{Kyriaki Kalimeri}, \bibinfo{person}{Jacopo
  Staiano}, \bibinfo{person}{Fabio Pianesi}, {and} \bibinfo{person}{Nicu
  Sebe}.} \bibinfo{year}{2012}\natexlab{}.
\newblock \showarticletitle{Connecting Meeting Behavior with Extraversion: A
  Systematic Study}.
\newblock \bibinfo{journal}{{\em IEEE Transactions on Affective Computing\/}}
  \bibinfo{volume}{3}, \bibinfo{number}{4} (\bibinfo{year}{2012}),
  \bibinfo{pages}{443--455}.
\newblock


\bibitem[\protect\citeauthoryear{Lovato, Perina, Sebe, Zandon\`a, Montagnini,
  Bicego, and Cristani}{Lovato et~al\mbox{.}}{2012}]%
        {Lovato2012}
\bibfield{author}{\bibinfo{person}{P. Lovato}, \bibinfo{person}{A. Perina},
  \bibinfo{person}{N. Sebe}, \bibinfo{person}{O. Zandon\`a},
  \bibinfo{person}{A. Montagnini}, \bibinfo{person}{M. Bicego}, {and}
  \bibinfo{person}{M. Cristani}.} \bibinfo{year}{2012}\natexlab{}.
\newblock \showarticletitle{Tell me what you like and {I}'ll tell you what you
  are: discriminating visual preferences on {Flickr} data}. In
  \bibinfo{booktitle}{{\em Proceedings of ACCV}}, Vol.~\bibinfo{volume}{7724}.
  \bibinfo{pages}{45--56}.
\newblock


\bibitem[\protect\citeauthoryear{Lowe}{Lowe}{2004}]%
        {lowe2004distinctive}
\bibfield{author}{\bibinfo{person}{David~G Lowe}.}
  \bibinfo{year}{2004}\natexlab{}.
\newblock \showarticletitle{Distinctive image features from scale-invariant
  keypoints}.
\newblock \bibinfo{journal}{{\em International Journal of Computer Vision\/}}
  \bibinfo{volume}{60}, \bibinfo{number}{2} (\bibinfo{year}{2004}),
  \bibinfo{pages}{91--110}.
\newblock


\bibitem[\protect\citeauthoryear{Machajdik and Hanbury}{Machajdik and
  Hanbury}{2010}]%
        {Machajdik2012}
\bibfield{author}{\bibinfo{person}{J. Machajdik} {and} \bibinfo{person}{A.
  Hanbury}.} \bibinfo{year}{2010}\natexlab{}.
\newblock \showarticletitle{Affective image classification using features
  inspired by psychology and art theory}. In \bibinfo{booktitle}{{\em
  Proceedings of the ACM-MM}}. \bibinfo{pages}{83--92}.
\newblock


\bibitem[\protect\citeauthoryear{Mardia and Jupp}{Mardia and Jupp}{2009}]%
        {mardia}
\bibfield{author}{\bibinfo{person}{K.V. Mardia} {and} \bibinfo{person}{P.E.
  Jupp}.} \bibinfo{year}{2009}\natexlab{}.
\newblock \bibinfo{booktitle}{{\em Directional Statistics}}.
\newblock \bibinfo{publisher}{Wiley}.
\newblock


\bibitem[\protect\citeauthoryear{Mohammadi and Vinciarelli}{Mohammadi and
  Vinciarelli}{2012}]%
        {mohammadi.vinciarelli12}
\bibfield{author}{\bibinfo{person}{Gelareh Mohammadi} {and}
  \bibinfo{person}{Alessandro Vinciarelli}.} \bibinfo{year}{2012}\natexlab{}.
\newblock \showarticletitle{Automatic personality perception: Prediction of
  trait attribution based on prosodic features}.
\newblock \bibinfo{journal}{{\em IEEE Transactions on Affective Computing\/}}
  \bibinfo{volume}{3}, \bibinfo{number}{3} (\bibinfo{year}{2012}),
  \bibinfo{pages}{273--284}.
\newblock


\bibitem[\protect\citeauthoryear{Musil, Preglej, Ropert, Klasinc, and
  Babic}{Musil et~al\mbox{.}}{2017}]%
        {musil.al.selfie2017}
\bibfield{author}{\bibinfo{person}{Bojan Musil}, \bibinfo{person}{Andrej
  Preglej}, \bibinfo{person}{Tadevž Ropert}, \bibinfo{person}{Lucia Klasinc},
  {and} \bibinfo{person}{Nenad~C. Babic}.} \bibinfo{year}{2017}\natexlab{}.
\newblock \showarticletitle{What Is Seen Is Who You Are: Are Cues in Selfie
  Pictures Related to Personality Characteristics?}
\newblock \bibinfo{journal}{{\em Frontiers in Psychology\/}}
  \bibinfo{volume}{8} (\bibinfo{year}{2017}), \bibinfo{pages}{82}.
\newblock
\showISSN{1664-1078}
\showDOI{%
\url{https://doi.org/10.3389/fpsyg.2017.00082}}


\bibitem[\protect\citeauthoryear{Oliva and Torralba}{Oliva and
  Torralba}{2001}]%
        {Oliva01}
\bibfield{author}{\bibinfo{person}{A. Oliva} {and} \bibinfo{person}{A.
  Torralba}.} \bibinfo{year}{2001}\natexlab{}.
\newblock \showarticletitle{Modeling the Shape of the Scene: A Holistic
  Representation of the Spatial Envelope}.
\newblock \bibinfo{journal}{{\em International Journal of Computer Vision\/}}
  \bibinfo{volume}{42}, \bibinfo{number}{3} (\bibinfo{year}{2001}),
  \bibinfo{pages}{145--175}.
\newblock


\bibitem[\protect\citeauthoryear{Pianesi, Mana, Cappelletti, Lepri, and
  Zancanaro}{Pianesi et~al\mbox{.}}{2008}]%
        {pianesi.al08}
\bibfield{author}{\bibinfo{person}{Fabio Pianesi}, \bibinfo{person}{Nadia
  Mana}, \bibinfo{person}{Alessandro Cappelletti}, \bibinfo{person}{Bruno
  Lepri}, {and} \bibinfo{person}{Massimo Zancanaro}.}
  \bibinfo{year}{2008}\natexlab{}.
\newblock \showarticletitle{Multimodal recognition of personality traits in
  social interactions}. In \bibinfo{booktitle}{{\em Proceedings of ICMI}}.
  \bibinfo{pages}{53--60}.
\newblock


\bibitem[\protect\citeauthoryear{Quercia, Kosinski, Stillwell, and
  Crowcroft}{Quercia et~al\mbox{.}}{2011}]%
        {quercia.al11a}
\bibfield{author}{\bibinfo{person}{Daniele Quercia}, \bibinfo{person}{Michal
  Kosinski}, \bibinfo{person}{David Stillwell}, {and} \bibinfo{person}{Jon
  Crowcroft}.} \bibinfo{year}{2011}\natexlab{}.
\newblock \showarticletitle{Our Twitter profiles, our selves: Predicting
  personality with Twitter}. In \bibinfo{booktitle}{{\em Proceedings of IEEE
  SocialCom}}. \bibinfo{pages}{180--185}.
\newblock


\bibitem[\protect\citeauthoryear{Quercia, Las~Casas, Pesce, Stillwell,
  Kosinski, Almeida, and Crowcroft}{Quercia et~al\mbox{.}}{2012}]%
        {quercia2012facebook}
\bibfield{author}{\bibinfo{person}{Daniele Quercia}, \bibinfo{person}{Diego~B
  Las~Casas}, \bibinfo{person}{Joao~Paulo Pesce}, \bibinfo{person}{David
  Stillwell}, \bibinfo{person}{Michal Kosinski}, \bibinfo{person}{Virgilio
  Almeida}, {and} \bibinfo{person}{Jon Crowcroft}.}
  \bibinfo{year}{2012}\natexlab{}.
\newblock \showarticletitle{Facebook and Privacy: The Balancing Act of
  Personality, Gender, and Relationship Currency.}. In \bibinfo{booktitle}{{\em
  Proceedings of AAAI ICWSM}}.
\newblock


\bibitem[\protect\citeauthoryear{Rammstedt and John}{Rammstedt and
  John}{2007}]%
        {rammstedt2007measuring}
\bibfield{author}{\bibinfo{person}{Beatrice Rammstedt} {and}
  \bibinfo{person}{Oliver~P John}.} \bibinfo{year}{2007}\natexlab{}.
\newblock \showarticletitle{Measuring personality in one minute or less: A
  10-item short version of the Big Five Inventory in English and German}.
\newblock \bibinfo{journal}{{\em Journal of research in Personality\/}}
  \bibinfo{volume}{41}, \bibinfo{number}{1} (\bibinfo{year}{2007}),
  \bibinfo{pages}{203--212}.
\newblock


\bibitem[\protect\citeauthoryear{Rosenberg and Egbert}{Rosenberg and
  Egbert}{2011}]%
        {rosenberg2011}
\bibfield{author}{\bibinfo{person}{J. Rosenberg} {and} \bibinfo{person}{N.
  Egbert}.} \bibinfo{year}{2011}\natexlab{}.
\newblock \showarticletitle{Online Impression Management: Personality Traits
  and Concerns for Secondary Goals as Predictors of Self-Presentation Tactics
  on Facebook}.
\newblock \bibinfo{journal}{{\em Journal of Computer-Mediated Communication\/}}
  \bibinfo{volume}{17}, \bibinfo{number}{1} (\bibinfo{year}{2011}),
  \bibinfo{pages}{106--116}.
\newblock


\bibitem[\protect\citeauthoryear{Schwartz, Eichstaedt, Kern, Dziurzynski,
  Ramones, Agrawal, Shah, Kosinski, Stillwell, Seligman,
  et~al\mbox{.}}{Schwartz et~al\mbox{.}}{2013}]%
        {schwartz2013personality}
\bibfield{author}{\bibinfo{person}{H~Andrew Schwartz},
  \bibinfo{person}{Johannes~C Eichstaedt}, \bibinfo{person}{Margaret~L Kern},
  \bibinfo{person}{Lukasz Dziurzynski}, \bibinfo{person}{Stephanie~M Ramones},
  \bibinfo{person}{Megha Agrawal}, \bibinfo{person}{Achal Shah},
  \bibinfo{person}{Michal Kosinski}, \bibinfo{person}{David Stillwell},
  \bibinfo{person}{Martin~EP Seligman}, {et~al\mbox{.}}}
  \bibinfo{year}{2013}\natexlab{}.
\newblock \showarticletitle{Personality, gender, and age in the language of
  social media: The open-vocabulary approach}.
\newblock \bibinfo{journal}{{\em PlOS One\/}} \bibinfo{volume}{8},
  \bibinfo{number}{9} (\bibinfo{year}{2013}), \bibinfo{pages}{e73791}.
\newblock


\bibitem[\protect\citeauthoryear{Segalin, Cheng, and Cristani}{Segalin
  et~al\mbox{.}}{2016}]%
        {segalin2016CNN}
\bibfield{author}{\bibinfo{person}{Cristina Segalin},
  \bibinfo{person}{Dong~Seon Cheng}, {and} \bibinfo{person}{Marco Cristani}.}
  \bibinfo{year}{2016}\natexlab{}.
\newblock \showarticletitle{Social profiling through image understanding:
  Personality inference using convolutional neural networks}.
\newblock \bibinfo{journal}{{\em Computer Vision and Image Understanding\/}}
  (\bibinfo{year}{2016}).
\newblock


\bibitem[\protect\citeauthoryear{Segalin, Perina, Cristani, and
  Vinciarelli}{Segalin et~al\mbox{.}}{2017}]%
        {segalin2016AffComp}
\bibfield{author}{\bibinfo{person}{Crisitina Segalin},
  \bibinfo{person}{Alessandro Perina}, \bibinfo{person}{Marco Cristani}, {and}
  \bibinfo{person}{Alessandro Vinciarelli}.} \bibinfo{year}{2017}\natexlab{}.
\newblock \showarticletitle{The pictures we like are our image: continuous
  mapping of favorite pictures into self-assessed and attributed personality
  traits}.
\newblock \bibinfo{journal}{{\em IEEE Transactions on Affective Computing\/}}
  \bibinfo{volume}{8}, \bibinfo{number}{2} (\bibinfo{year}{2017}),
  \bibinfo{pages}{268--285}.
\newblock


\bibitem[\protect\citeauthoryear{Sorokowski, Sorokowska, Oleszkiewicz,
  Frackowiak, Huk, and Pisanski}{Sorokowski et~al\mbox{.}}{2015}]%
        {sorokowski2015selfie}
\bibfield{author}{\bibinfo{person}{P Sorokowski}, \bibinfo{person}{A
  Sorokowska}, \bibinfo{person}{A Oleszkiewicz}, \bibinfo{person}{T
  Frackowiak}, \bibinfo{person}{A Huk}, {and} \bibinfo{person}{K Pisanski}.}
  \bibinfo{year}{2015}\natexlab{}.
\newblock \showarticletitle{Selfie posting behaviors are associated with
  narcissism among men}.
\newblock \bibinfo{journal}{{\em Personality and Individual Differences\/}}
  \bibinfo{volume}{85} (\bibinfo{year}{2015}), \bibinfo{pages}{123--127}.
\newblock


\bibitem[\protect\citeauthoryear{Staiano, Lepri, Aharony, Pianesi, Sebe, and
  Pentland}{Staiano et~al\mbox{.}}{2012}]%
        {staiano.al12}
\bibfield{author}{\bibinfo{person}{Jacopo Staiano}, \bibinfo{person}{Bruno
  Lepri}, \bibinfo{person}{Nadav Aharony}, \bibinfo{person}{Fabio Pianesi},
  \bibinfo{person}{Nicu Sebe}, {and} \bibinfo{person}{Alex Pentland}.}
  \bibinfo{year}{2012}\natexlab{}.
\newblock \showarticletitle{Friends don't lie: inferring personality traits
  from social network structure}. In \bibinfo{booktitle}{{\em Proceedings of
  Ubicomp}}. \bibinfo{pages}{321--330}.
\newblock


\bibitem[\protect\citeauthoryear{Stillwell and Kosinski}{Stillwell and
  Kosinski}{2012}]%
        {stillwell2012mypersonality}
\bibfield{author}{\bibinfo{person}{David~J Stillwell} {and}
  \bibinfo{person}{Michal Kosinski}.} \bibinfo{year}{2012}\natexlab{}.
\newblock \showarticletitle{myPersonality project: Example of successful
  utilization of online social networks for large-scale social research}.
\newblock \bibinfo{journal}{{\em Proceedings of MobiSys\/}}
  (\bibinfo{year}{2012}).
\newblock


\bibitem[\protect\citeauthoryear{Tamura, Mori, and Yamawaki}{Tamura
  et~al\mbox{.}}{1978}]%
        {tamura}
\bibfield{author}{\bibinfo{person}{H. Tamura}, \bibinfo{person}{S. Mori}, {and}
  \bibinfo{person}{T. Yamawaki}.} \bibinfo{year}{1978}\natexlab{}.
\newblock \showarticletitle{Texture features corresponding to visual
  perception}.
\newblock \bibinfo{journal}{{\em IEEE Trans. on Systems, Man and
  Cybernetics\/}} \bibinfo{volume}{8}, \bibinfo{number}{6}
  (\bibinfo{year}{1978}), \bibinfo{pages}{460--473}.
\newblock


\bibitem[\protect\citeauthoryear{Tettegah}{Tettegah}{2016}]%
        {tettegah2016emotions}
\bibfield{author}{\bibinfo{person}{Sharon Tettegah}.}
  \bibinfo{year}{2016}\natexlab{}.
\newblock \bibinfo{booktitle}{{\em Emotions, Technology, and Social Media}}.
\newblock \bibinfo{publisher}{Academic Press}.
\newblock


\bibitem[\protect\citeauthoryear{Utz}{Utz}{2010}]%
        {Utz2010}
\bibfield{author}{\bibinfo{person}{S. Utz}.} \bibinfo{year}{2010}\natexlab{}.
\newblock \showarticletitle{Show me your friends and I will tell you what type
  of person you are: How one's profile, number of friends, and type of friends
  influence impression formation on social network sites}.
\newblock \bibinfo{journal}{{\em Journal of Computer-Mediated Communication\/}}
  \bibinfo{volume}{15}, \bibinfo{number}{2} (\bibinfo{year}{2010}),
  \bibinfo{pages}{314--335}.
\newblock


\bibitem[\protect\citeauthoryear{Van Der~Heide, D'Angelo, and Schumaker}{Van
  Der~Heide et~al\mbox{.}}{2012}]%
        {van2012effects}
\bibfield{author}{\bibinfo{person}{Brandon Van Der~Heide},
  \bibinfo{person}{Jonathan~D D'Angelo}, {and} \bibinfo{person}{Erin~M
  Schumaker}.} \bibinfo{year}{2012}\natexlab{}.
\newblock \showarticletitle{The effects of verbal versus photographic
  self-presentation on impression formation in Facebook}.
\newblock \bibinfo{journal}{{\em Journal of Communication\/}}
  \bibinfo{volume}{62}, \bibinfo{number}{1} (\bibinfo{year}{2012}),
  \bibinfo{pages}{98--116}.
\newblock


\bibitem[\protect\citeauthoryear{Vazire and Carlson}{Vazire and
  Carlson}{2011}]%
        {vazire2011}
\bibfield{author}{\bibinfo{person}{S. Vazire} {and} \bibinfo{person}{E.N.
  Carlson}.} \bibinfo{year}{2011}\natexlab{}.
\newblock \showarticletitle{Others sometimes know us better than we know
  ourselves}.
\newblock \bibinfo{journal}{{\em Current Directions in Psychological
  Science\/}} \bibinfo{volume}{20}, \bibinfo{number}{2} (\bibinfo{year}{2011}),
  \bibinfo{pages}{104--108}.
\newblock


\bibitem[\protect\citeauthoryear{Vazire and Gosling}{Vazire and
  Gosling}{2004}]%
        {vazire.gosling04}
\bibfield{author}{\bibinfo{person}{Simine Vazire} {and}
  \bibinfo{person}{Samuel~D Gosling}.} \bibinfo{year}{2004}\natexlab{}.
\newblock \showarticletitle{e-Perceptions: personality impressions based on
  personal websites}.
\newblock \bibinfo{journal}{{\em Journal of Personality and Social
  Psychology\/}} \bibinfo{volume}{87}, \bibinfo{number}{1}
  (\bibinfo{year}{2004}), \bibinfo{pages}{123}.
\newblock


\bibitem[\protect\citeauthoryear{Vedaldi}{Vedaldi}{2007}]%
        {Vedaldi07anopen}
\bibfield{author}{\bibinfo{person}{Andrea Vedaldi}.}
  \bibinfo{year}{2007}\natexlab{}.
\newblock \bibinfo{title}{An open implementation of the {SIFT} detector and
  descriptor}.
\newblock   (\bibinfo{year}{2007}).
\newblock


\bibitem[\protect\citeauthoryear{Vinciarelli and Mohammadi}{Vinciarelli and
  Mohammadi}{2014}]%
        {vinciarelli2014survey}
\bibfield{author}{\bibinfo{person}{Alessandro Vinciarelli} {and}
  \bibinfo{person}{Gelareh Mohammadi}.} \bibinfo{year}{2014}\natexlab{}.
\newblock \showarticletitle{A survey of personality computing}.
\newblock \bibinfo{journal}{{\em IEEE Transactions on Affective Computing\/}}
  \bibinfo{volume}{5}, \bibinfo{number}{3} (\bibinfo{year}{2014}),
  \bibinfo{pages}{273--291}.
\newblock


\bibitem[\protect\citeauthoryear{Viola and Jones}{Viola and Jones}{2001}]%
        {Viola01}
\bibfield{author}{\bibinfo{person}{P. Viola} {and} \bibinfo{person}{M. Jones}.}
  \bibinfo{year}{2001}\natexlab{}.
\newblock \showarticletitle{Rapid object detection using a boosted cascade of
  simple features}. In \bibinfo{booktitle}{{\em Proceedings of IEEE CVPR}}.
  \bibinfo{pages}{511--518}.
\newblock


\bibitem[\protect\citeauthoryear{Wu, Chang, and Yuan}{Wu et~al\mbox{.}}{2014}]%
        {wu2014facebook}
\bibfield{author}{\bibinfo{person}{Yen-Chun~Jim Wu}, \bibinfo{person}{Wei-Hung
  Chang}, {and} \bibinfo{person}{Chih-Hung Yuan}.}
  \bibinfo{year}{2014}\natexlab{}.
\newblock \showarticletitle{Do Facebook profile pictures reflect user's
  personality?}
\newblock \bibinfo{journal}{{\em Computers in Human Behavior\/}}
  \bibinfo{volume}{51} (\bibinfo{year}{2014}), \bibinfo{pages}{880--889}.
\newblock


\bibitem[\protect\citeauthoryear{Youyou, Kosinski, and Stillwell}{Youyou
  et~al\mbox{.}}{2015}]%
        {youyou.al15}
\bibfield{author}{\bibinfo{person}{Wu Youyou}, \bibinfo{person}{Michal
  Kosinski}, {and} \bibinfo{person}{David Stillwell}.}
  \bibinfo{year}{2015}\natexlab{}.
\newblock \showarticletitle{Computer-based personality judgments are more
  accurate than those made by humans}.
\newblock \bibinfo{journal}{{\em PNAS\/}} (\bibinfo{year}{2015}),
  \bibinfo{pages}{1--5}.
\newblock


\end{thebibliography}

\end{document}